\documentclass[journal]{IEEEtran}
\usepackage{amsmath,amsfonts}
\usepackage{algorithmic}
\usepackage{array}
\usepackage[caption=false,font=normalsize,labelfont=sf,textfont=sf]{subfig}
\usepackage{textcomp}
\usepackage{stfloats}
\usepackage{url}
\usepackage{verbatim}
\usepackage{graphicx}
\usepackage{booktabs}
\usepackage{multirow}
\usepackage{bookmark}

\def\BibTeX{{\rm B\kern-.05em{\sc i\kern-.025em b}\kern-.08em
    T\kern-.1667em\lower.7ex\hbox{E}\kern-.125emX}}
\usepackage{balance}

\begin{document}

\title{SwinNet: Swin Transformer Drives Edge-Aware RGB-D and RGB-T Salient Object Detection}
\author{Zhengyi Liu*, Yacheng Tan, Qian He and Yun Xiao
\thanks{This work was supported by National Natural Science Foundation of China (62006002) and Natural Science Foundation of Anhui Province (1908085MF182).(Corresponding author: Zhengyi Liu)}
\thanks{Zhengyi Liu, Yacheng Tan and Qian He are with Key Laboratory of Intelligent Computing and Signal Processing of Ministry of Education, School of Computer Science and Technology, Anhui University, Hefei, China(e-mail: liuzywen@ahu.edu.cn,1084043983@qq.com,1819469871@qq.com).}
\thanks{Yun Xiao is with Key Laboratory of Intelligent Computing and Signal Processing of Ministry of Education, School of Artificial Intelligence, Anhui University, Hefei, China(e-mail: 280240406@qq.com).}
\thanks{Copyright $\copyright$2021 IEEE. Personal use of this material is permitted. However, permission to use this material for any other purposes must be obtained from the IEEE by sending an email to pubs-permissions@ieee.org.}
}

\markboth{Journal of \LaTeX\ Class Files,~Vol.~18, No.~9, November~2021}%
{SwinNet: Swin Transformer drives edge-aware RGB-D and RGB-T salient object detection}

\maketitle

\begin{abstract}
Convolutional neural networks (CNNs) are good at extracting contexture features within certain receptive fields, while transformers can model the global long-range dependency features. By absorbing the advantage of transformer and the merit of CNN, Swin Transformer shows strong feature representation ability. Based on it, we propose a cross-modality fusion model, \textbf{SwinNet}, for RGB-D and RGB-T salient object detection.
It is driven by Swin Transformer to extract the hierarchical features, boosted by attention mechanism to bridge the gap between two modalities, and guided by edge information to sharp the contour of salient object.
To be specific, two-stream Swin Transformer encoder first  extracts multi-modality features, and then spatial alignment and channel re-calibration module is presented to optimize intra-level cross-modality features. To clarify the fuzzy boundary, edge-guided decoder achieves inter-level cross-modality fusion under the guidance of edge features. The proposed model outperforms the state-of-the-art models on RGB-D and RGB-T datasets, showing that it provides more insight into the cross-modality complementarity task.\href{https://github.com/liuzywen/SwinNet}{https://github.com/liuzywen/SwinNet}
\end{abstract}


\section{Introduction}
Salient object detection (SOD) simulates the visual attention mechanism to capture the prominent object. As described in the SOD review\cite{cong2018review}, SOD has been extended from RGB image\cite{hu2020sac,tu2020edge,wang2020deep} to RGB-D image\cite{liu2019salient,liu2021visual}, a group of images\cite{gao2020co,han2017unified} and video\cite{guo2019motion,xu2019video}. Recently, SOD in
RGB-T image\cite{zhou2021ecffnet}, light field image\cite{zhang2020multi,piao2020dut,zhang2021cma}, high-resolution image\cite{zhang2021looking,zeng2019towards}, optical remote sensing image\cite{li2020parallel,li2019nested,zhang2020dense} and 360$^{\circ}$ omnidirectional image\cite{ma2020stage,huang2020fanet} have been gradually researched.
SOD can benefit many image and video processing tasks, such as image
segmentation~\cite{yarlagadda2021saliency,huang2021graph}, tracking~\cite{ma2017saliency,hong2015online,zhang2020non}, retrieval~\cite{gao2015database}, compression~\cite{ji2013video}, cropping~\cite{wang2018deep,xu2021saliency}, retargeting\cite{ahmadi2021context},
quality assessment~\cite{jiang2017optimizing} and activity prediction\cite{weng2021human}.

When the light is insufficient or the background is cluttered in a scene, SOD is still a challenge issue.
With the widespread use of depth cameras and infrared imaging devices, the depth or thermal infrared information as the supplementary modality has shown the advantages to SOD performance improvements, because
the depth image can provide the more geometry information and the thermal image can capture the radiated heat of objects especially under adverse weather and lighting conditions.
Nevertheless, how to effectively implement cross-modality information fusion is still challenging, which can significantly affect the achievement of robust performance.

In the past few years, convolutional neural networks (CNNs) have achieved milestones in RGB-D and RGB-T SOD.
However, CNN gathers information from neighborhood pixels and loses spatial information due to pooling operation.
It is not easy to learn global long-range semantic information interaction well.
Recently, Swin Transformer\cite{liu2021swin} is proposed.
It implements pairwise entity interactions within a local window by multi-head self-attention, and establishes long-range dependency across windows by shifted windowing scheme.
Features extracted from transformer have more global attributes than those from CNN.
By absorbing the locality, translation invariance and hierarchical merits of CNN, Swin Transformer can be used as backbone network to extract hierarchical information of each modality.
The features from different modalities show the different attribution.
They consistently display the common salient position in the spatial aspect, and respectively show the different salient content in the channel aspect,
so spatial alignment and channel re-calibration module is designed to boost the extracted features based on attention mechanism.
In addition, SOD task is essentially a pixel-level dense prediction task. After the feature extraction of encoder, the multi-level features with different receptive field and spatial resolution need to be progressively combined  by upsampling and skip connection. In the decoding process, shallow-level features exhibit the detailed boundary information, and meanwhile it also brings some background noises.
Therefore, edge-aware module is presented to extract the edge feature, and further to guide the decoder for both suppressing the shallow-layer noise and refining the contour of objects.

Our main contributions can be summarized as follows:
\begin{itemize}
    \item
    A novel SOD model (SwinNet) for both RGB-D and RGB-T tasks  built upon the Swin Transformer backbone is proposed. It extracts  discriminative features from Swin Transformer backbone which absorbs the local advantage of convolution neural network and the long-range dependency merit of transformer, outperforming the state-of-the-art (SOTA) RGB-D and RGB-T SOD models.
    \item A newly designed spatial alignment and channel re-calibration module is used to optimize the features of each modality based on attention mechanism, achieving intra-layer cross-modality fusion from the spatial and channel aspects.
    \item The proposed edge-guided decoder achieves inter-layer cross-modal fusion under the guidance of edge-aware module, generating the sharper contour.
\end{itemize}
\section{Related works}
\subsection{RGB-D salient object detection}
Salient object detection has achieved the massive improvement by combining the other modality with color modality.
Depth image is exactly a good supplement, because it provides more reliable spatial structure information and insensitive to the variations of the environment lights and colors.

Cong et al.\cite{cong2019going} introduces depth information in the initialization, refinement and optimization of saliency map, achieving transfer from existing RGB SOD models to RGB-D SOD models.
In these years, 
attention mechanism\cite{li2020asif,liu2020learning1,liu2020learning,fan2020bbs,zhang2021bts,chen2020dpanet}, multi-task learning\cite{wu2021multiscale,zhang2020select,ji2020accurate,li2020rgb}, knowledge distillation\cite{piao2020a2dele}, graph neural networks\cite{luo2020cascade}, neural architecture search\cite{Sun2021DeepRS}, 3D convolutional neural networks\cite{chen2021rd3d}, self-supervised learning\cite{zhao2021self}, generative adversarial networks\cite{jiang2020cmsalgan}, disentanglement and reconstruction\cite{chen2020rgbd} are applied to solve SOD task.
The intrinsic defect of CNN limits above methods in learning global long-range dependencies.
Visual Saliency Transformer (VST)\cite{liu2021visual} 
propagates long-range dependencies across modalities by Scaled Dot-Product Attention\cite{NIPS2017_3f5ee243} between the queries from one modality with the keys of the other modality.
It also designs reverse T2T to decode and introduces edge detection to improve the performance.
Motivated by its success, we introduce Swin Transformer as backbone to enhance the feature representation, and then use transformer encoder and CNN decoder to complete the SOD task.

\subsection{RGB-T salient object detection}
The thermal image can capture the radiated heat of objects, and it is insensitive to lighting and weather conditions, and suitable for handling scenes captured under adverse conditions, for example, total darkness environment, foggy weather, and cluttered backgrounds.
Therefore, thermal image is a promising supplement to the RGB image for SOD.
In earlier years,
RGB-T SOD  adopts machine learning methods, for example,
SVM\cite{ma2017learning}, ranking models\cite{tu2019m3s,wang2018rgb, tang2019rgbt} and graph learning\cite{tu2019rgb}.
With the development of CNN, Tu et al.\cite{tu2020rgbt} propose a baseline model which combines CNN with attention mechanism. Zhang et al.\cite{zhang2019rgb,zhang2020revisiting} propose
two end-to-end CNN based RGB-T SOD models  to achieve multi-scale, multi-modality and multi-level fusion.
ECFFNet\cite{zhou2021ecffnet} achieves more effectively cross-modality fusion, and enhances salient object boundaries by a bilateral reversal fusion of foreground and background information.
MIDD\cite{tu2021multiinteractive} proposes multi-interactive dual-decoder to integrate the multi-level interactions of dual modalities and global contexts.
MMNet\cite{gao2021unified} simulates visual color stage doctrine to fuse cross-modal features in stages, and designs bi-directional multi-scale decoder to capture both local and global information.
CGFNet\cite{wang2021cgfnet} adopts the guidance manner of one modality on the other modality to fuse two modalities.
CSRNet\cite{huo2021efficient} uses context-guided cross modality fusion module to fuse two modalities, and designs a stacked refinement network to refine the segmentation results.
Pushed by the global merit of Transformer in computer vision, we propose transformer based method to detect the salient object in RGB-T images.

\subsection{Transformer}
Vaswani et al.\cite{NIPS2017_3f5ee243} first proposes  transformer
with stacked multi-head self-attention  and point-wise feed-forward layers in machine translation task.
Recently, inspired by successful ViT\cite{yuan2021tokens}, transformer variants emerge explosively.
T2T\cite{yuan2021tokens} progressively structurizes the image to tokens by recursively aggregating neighboring tokens into one token.
CvT\cite{wu2021cvt} adds convolutional layers into the multi-head self-attention.
PVT\cite{wang2021pyramid} introduces a progressive shrinking pyramid to
reduce the sequence length of the transformer.
DPT\cite{ranftl2021vision} assembles tokens from multiple stages of the vision transformer and progressively combines them into full-resolution predictions using a convolutional decoder.
Swin transformer\cite{liu2021swin} designs the shifted window-based multi-head attentions to reduce the computation cost.
CAT\cite{lin2021cat} alternately applies attention inner patch and between patches to maintain the performance with lower computational cost and builds a cross attention hierarchical network.
Due to the perfect performance of Swin Transformer, it is used as the backbone network.
\section{Proposed method}
\subsection{Overview}
\begin{figure*}[!htp]

  \includegraphics[width=\textwidth]{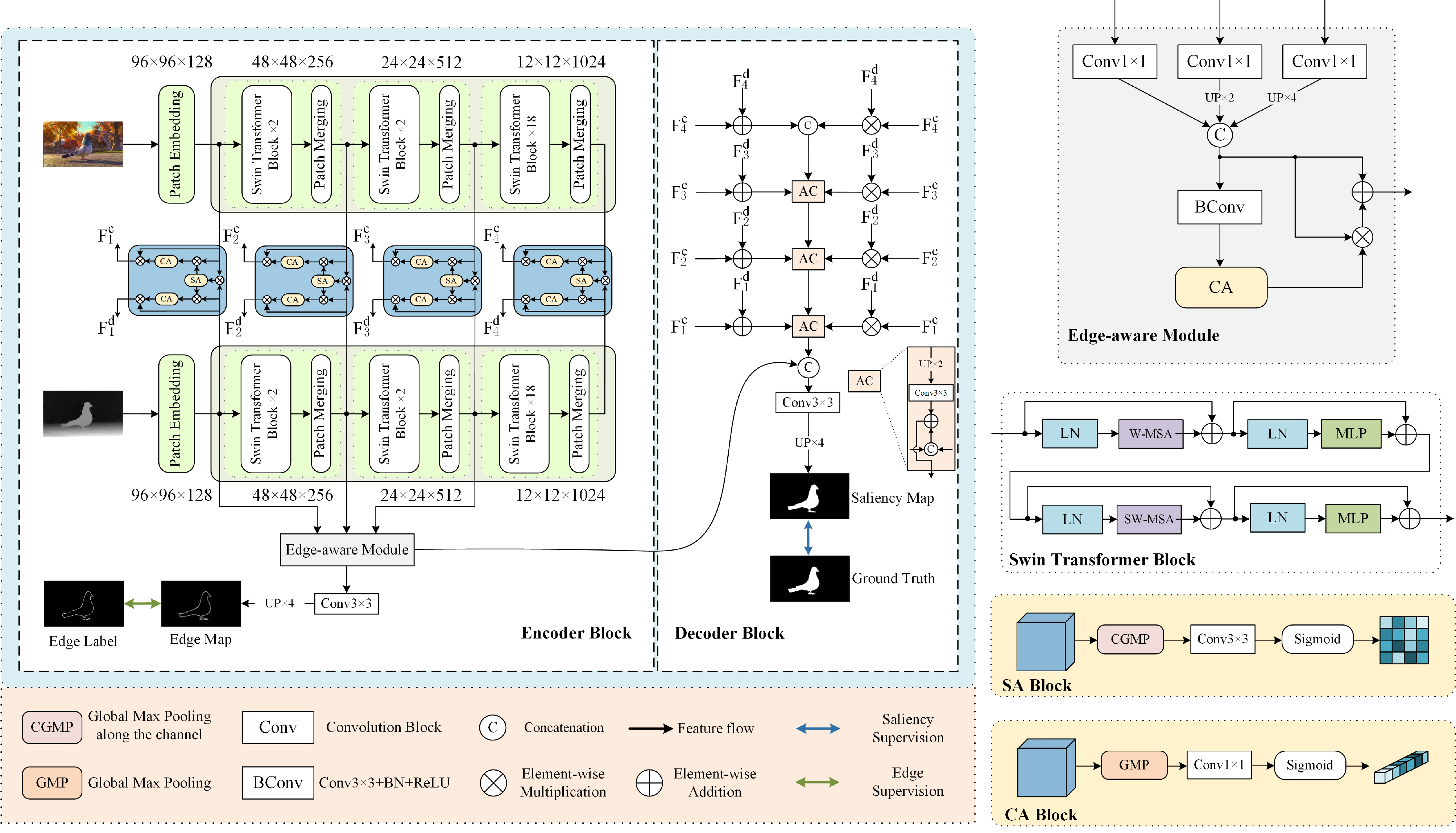}
  \caption{An overview of our proposed SwinNet. It consists of a two-stream backbone, a spatial alignment and channel re-calibration module, an edge-aware module and an edge-guided decoder. Multi-modal hierarchical features from two-stream backbone will be fed into spatial alignment and channel re-calibration modules to generate the enhanced features $F_i^c$ and $F_i^d$ (i=1,$\cdots$, 4). Besides, edge feature is generated from edge-aware module which process the shallow-layer features of the depth backbone. At last, in the edge-guided decoder, enhanced features and edge feature are combined to generate saliency map.}
  \label{fig:main}
\end{figure*}

The overall framework of the proposed model is illustrated in Fig.\ref{fig:main}, which consists of a two-stream backbone, a spatial alignment and channel re-calibration module, an edge-aware module and an edge-guided decoder.  Note that since RGB-D and RGB-T SOD are the same multi-modality fusion tasks, for brevity, below we only elaborate the implementation detail of RGB-D SOD task, because that of RGB-T is the same.

\subsection{Two-stream Swin Transformer backbone}
Swin Transformer has the flexibility to model at various
scales and has linear computational complexity with
respect to image size\cite{liu2021swin}. We adopt two Swin Transformers to extract hierarchical features from multi-modality image pairs. Considering the complexity and efficiency, Swin-B version\cite{liu2021swin} is adopted.

Each Swin Transformer first splits the input single-modality image into non-overlapping patches by a patch embedding. The feature of each patch in color stream is set as a concatenation of the raw pixel RGB values, while that in depth stream is set as a concatenation of three copied depth values. Then, they are fed into the multi-stage feature transformation.
With the increasing depth of the network, the number of tokens is gradually reduced by patch merging layers  to produce the hierarchical representation of each modality, which can be denoted as $\{ST_i^c\}_{i=1}^4$ and $\{ST_i^d\}_{i=1}^4$, respectively.
\subsection{Spatial  alignment and channel re-calibration module}
On one hand, since the position of salient objects in multi-modality image pairs should be the same, the features from different modalities need to be aligned at first to show the common salient position. On the other hand, since
RGB image shows  more appearance and texture information, and depth image exhibits more spatial cue,  the features from different modalities are different in the importance of feature channels, and the multi-modality features need to be re-calibrated to emphasize their respective salient content. Therefore, the spatial  alignment and channel re-calibration module is proposed. It first aligns two modalities in spatial part, and then recalibrates respective channel part to pay more attention to the salient content in each modality.

Specifically, given the color features $ST_i^c$ and depth feature $ST_i^d$ at a certain hierarchy $i\in\{1,\cdots,4\}$, we first compute their common spatial attention map $SA_i$ as:
\begin{small}
\begin{equation}
\begin{aligned}
SA_i=SA(ST_i^c \times ST_i^d)
\end{aligned}
\end{equation}
\end{small}\\
where ``$\times$" means element-wise multiplication operation, and $SA(\cdot)$ denotes spatial attention operation which is defined as:
\begin{small}
\begin{equation}
\begin{aligned}
SA(x)=Sigmoid(Conv_3(CGMP(x)))
\end{aligned}
\end{equation}
\end{small}\\
where $CGMP(\cdot)$ means global max pooling operation along channel direction, $Conv_3(\cdot)$ represents the convolution operation with the kernel size 3$\times$3, and $Sigmoid(\cdot)$ denotes the sigmoid activation function.

Next, the common spatial attention map is served as the weight of color feature and depth feature to achieve the spatial alignment of both modalities by:
\begin{small}
\begin{equation}
\begin{aligned}
ST1_i^c=SA_i \times ST_i^c\\
ST1_i^d=SA_i \times ST_i^d
\end{aligned}
\end{equation}
\end{small}

Third, the aligned features in spatial part $ST1_i^l(l\in\{c,d\})$ are performed channel attention respectively, to generate channel attention map which shows more weights on the more salient content in each modality by:
\begin{small}
\begin{equation}
\begin{aligned}
CA_i^c=CA(ST1_i^c)\\
CA_i^d=CA(ST1_i^d)
\end{aligned}
\end{equation}
\end{small}\\
where $CA(\cdot)$ denotes channel attention operation which is defined as:
\begin{small}
\begin{equation}
\begin{aligned}
CA(x)=Sigmoid(Conv_1(GMP(x)))
\end{aligned}
\end{equation}
\end{small}\\
where $GMP(\cdot)$ means the global max pooling operation, $Conv_1$ represents the convolution operation with the kernel size 1$\times$1.

Last, each channel attention map is multiplied with original feature to achieve the channel re-calibration.
\begin{small}
\begin{equation}
\begin{aligned}
F_i^c=CA_i^c \times ST_i^c\\
F_i^d=CA_i^d \times ST_i^d
\end{aligned}
\end{equation}
\end{small}

After the spatial alignment and channel re-calibration module, the enhanced features $F_i^l(l\in \{c,d\}$) achieve the position alignment and channel re-calibration, which show the stronger representation ability.
\subsection{Edge-aware module}
As we all known, high-layer features express more semantic information, while shallow-layer features carry more details. Meanwhile, salient objects are more likely to exhibit pop-out structure in the depth image\cite{feng2016local}. It is easy to depict the object contours by depth contrast.
Therefore, the shallow-layer features of the depth backbone are used to produce the edge feature.

Specifically, $ST^d_i(i=1,2,3)$ are performed 1$\times$1 convolutional operation and upsampling operation to generate three features with the same size, and then they are concatenated to generate edge feature.
\begin{small}
\begin{equation}
\begin{aligned}
F_e=Concat(Conv_1(ST_1^d),Up_2(Conv_1(ST_2^d)),Up_4(Conv_1(St_3^d)))
\end{aligned}
\end{equation}
\end{small}\\
where
$Up_x(\cdot)$ denotes $x\times$upsampling operation, and $Concat(\cdot)$ means the concatenation operation.

Next, the obtained edge feature is performed a channel attention and a residual connection to generate the clearer edge information by:
\begin{small}
\begin{equation}
\begin{aligned}
F'_e=F_e\times CA(BConv(F_e))+F_e
\end{aligned}
\end{equation}
\end{small}\\
where $BConv(\cdot)$ represents convolutional operation with kernel size 3$\times$3 followed by a batch normalization layer and a ReLU activation function, and ``+" means element-wise addition operation.

The edge-aware module outputs the edge features $F'_e$  which will be used to guide the decoding process of the model and enhance the details.

\subsection{Edge-guided decoder}
After spatial alignment and channel re-calibration and edge feature extraction, decoder combines the enhanced hierarchical features of different modalities with the edge features to produce the edge-guided salient feature.

Specifically, the aligned and re-calibrated color and depth features from two modalities $F_i^c$ and $F_i^d$ at a certain hierarchy $i\in\{1,\cdots,4\}$ are fused by the addition, multiplication and concatenation operation by:
\begin{small}
\begin{equation}
\begin{aligned}
F_i=Concat((F_i^d + F_i^c), (F_i^d \times F_i^c))
\end{aligned}
\end{equation}
\end{small}

Next, according to the decoding idea widely used in U-Net framework\cite{ronneberger2015u}, the high-level fused feature is progressively aggregated into the shallow-layer fused features by:
\begin{small}
\begin{equation}
\begin{aligned}
FF_i= \left\{\begin{matrix}
	F_i+Conv_3(Up_2(FF_{i+1})),&i=1,2,3 \\
	F_i,&i=4
	\end{matrix}\right.
\end{aligned}
\end{equation}
\end{small}

At last, edge feature from edge-aware module is combined with fused feature to generate the edge-guided salient feature $F_{s}$.
\begin{small}
\begin{equation}
\begin{aligned}
F_{s}=Concat(F'_e,FF_1)
\end{aligned}
\end{equation}
\end{small}
\subsection{Loss function}
The loss function $L$ is defined as:
\begin{small}
\begin{equation}
\begin{aligned}
   L =L_{e}(S_e)+L_{s}(S)
\end{aligned}
\end{equation}
\end{small}\\
where $L_{e}$ and $L_{s}$ denote edge loss  and saliency loss, respectively.
\subsubsection{Edge loss}
The edge map is generated from edge-aware module. Specifically, edge feature $F'_e$ is fed into a convolution layer and a upsampling layer to generate edge map $S_e$ by:
\begin{small}
\begin{equation}
\begin{aligned}
    S_e=Up_4(Conv_3(F'_e))
\end{aligned}
\end{equation}
\end{small}

The edge ground truth  can be easily got from saliency map ground truth by Canny edge detector~\cite{canny1986computational}. It is used to supervise the edge map $S_e$.
The edge loss $L_{e}$   adopts the cross-entropy loss, and it is defined as:
\begin{small}
\begin{equation}
\begin{aligned}
    L_{e}(S_e)=-\sum_{j\in Z_+} log  Pr(y_j=1|S_{e})
    -\sum_{j\in Z_-} log  Pr(y_j=0|S_{e})
\end{aligned}
\end{equation}
\end{small}\\
where $Z_+$ and $Z_{-}$ denote the edge pixels set and background pixels set respectively. $Pr(y_j=1|S_{e})$ is the prediction map  in which each value denotes the edge confidence for the pixel.

\subsubsection{Saliency loss}
The final saliency map can be generated from edge-guided decoder. Specifically, the edge-guided salient feature $F_s$ is fed into a convolution layer and a upsampling layer to generate saliency map $S$ by:
\begin{small}
\begin{equation}
\begin{aligned}
    S=Up_4(Conv_3(F_s))
\end{aligned}
\end{equation}
\end{small}

The saliency loss $L_{s}$   adopts the cross-entropy loss, and it is defined as:
\begin{small}
\begin{equation}
\begin{aligned}
    L_{s}(S)=-\sum_{j\in Y_+} log  Pr(y_j=1|S)
    -\sum_{j\in Y_-} log  Pr(y_j=0|S)
\end{aligned}
\end{equation}
\end{small}\\
where $Y_+$ and $Y_{-}$ denote the salient region pixels set and non-salient pixels set respectively. $Pr(y_j=1|S)$ is the prediction map in which each value denotes the salient region confidence for the pixel.

\section{Experiments}

\subsection{Datasets and evaluation metrics}

\subsubsection{Datasets}
For RGB-D SOD, we evaluate the proposed method on several challenging RGB-D SOD datasets.
NLPR~\cite{peng2014rgbd} includes  1,000 images with single or multiple salient objects.
NJU2K~\cite{ju2014depth} consists of 2,003 stereo image pairs and ground-truth maps with different objects, complex and challenging scenes.
STERE~\cite{niu2012leveraging} incorporates 1,000 pairs of binocular images downloaded from the Internet.
DES~\cite{cheng2014depth} has 135 indoor images collected by
Microsoft Kinect.
SIP~\cite{fan2020rethinking} contains 1,000 high-resolution images of multiple salient persons.
DUT~\cite{piao2019depth} contains 1,200 images captured by Lytro camera in real life scenes.
For the sake of fair comparison, we use the same training dataset as in~\cite{fan2020rethinking,chen2020progressively}, which consists of 1,485 images from the NJU2K dataset and 700 images from the NLPR dataset. The remaining images are used for testing.
In addition, on the DUT dataset, we follow
the same protocols as in ~\cite{piao2019depth, zhao2020single,piao2020a2dele,li2020rgb,ji2020accurate} to add additional 800 pairs from DUT for training and test on the
remaining 400 pairs.
In summary, our training set contains 2,185 paired RGB and depth images, but when testing is conducted on DUT, our training set contains 2,985 paired ones.

For RGB-T SOD, we evaluate the proposed method on three RGB-T SOD datasets.
VT821\cite{wang2018rgb} contains 821 manually registered image pairs.
VT1000\cite{tu2019rgb} contains 1,000 RGB-T image pairs captured with highly aligned RGB and thermal cameras.
VT5000\cite{tu2020rgbt} contains 5,000 pairs of high-resolution, high-diversity and low-deviation RGB-T images.
For the sake of fair comparison, we use the same training dataset as in~\cite{tu2021multiinteractive,tu2020multi,zhou2021ecffnet}, which consists of 2,500 image pairs in VT5000. The rest image pairs are used for testing.

\subsubsection{Evaluation Metrics}
We adopt  widely used metrics to evaluate the performance of our model and SOTA RGB-D and RGB-T SOD models.
They are the precision-recall (PR) curve~\cite{borji2015salient}, S-measure~\cite{fan2017structure},
F-measure~\cite{achanta2009frequency}, E-measure~\cite{fan2018enhanced} and mean absolute error (MAE)~\cite{perazzi2012saliency}. Specifically, the PR curve plots precision and recall values by  setting a series of thresholds on the saliency maps to get the binary masks and further comparing them with the ground truth maps.
The S-measure can evaluate both region-aware and object-aware structural similarity between saliency map and ground truth.
The F-measure is the weighted harmonic mean of precision and recall, which can evaluate the overall performance.
The E-measure simultaneously captures global statistics and local pixel matching information.
The MAE measures the average of the per-pixel absolute difference between the saliency maps and the ground truth maps. In our experiment, E-measure and F-measure adopt the adaptive values.

\subsection{Implementation details}
During the training and testing phase, the input RGB, depth and thermal images are resized to 384$\times$384. Since the depth image is single-channel data, it is copied to form three-channel image which is the same as RGB and thermal images. Multiple enhancement strategies are used for all training images, i.e., random flipping, rotating and border clipping. Parameters of the backbone network are initialized with the pretrained  parameters of Swin-B network\cite{liu2021swin}. The rest of parameters are initialized to PyTorch default settings. We employ the Adam optimizer~\cite{kingma2014adam} to train our network with a batch size of 3 and an initial learning rate 5e-5, and the learning rate will be divided by
10 every 100 epochs. Our model is trained on a machine with a single NVIDIA RTX 2080Ti GPU. The model converges within 200 epochs, which takes nearly 26 hours.

\subsection{Comparisons with SOTAs}
\subsubsection{RGB-D SOD}
For RGB-D SOD, our model is compared with several SOTA RGB-D SOD algorithms, including D3Net~\cite{fan2020rethinking}, ASIF-Net\cite{li2020asif}, ICNet~\cite{li2020icnet}, DCMF~\cite{chen2020rgbd}, DRLF~\cite{wang2020data}, SSF~\cite{zhang2020select}, SSMA~\cite{liu2020learning}, A2dele~\cite{piao2020a2dele}, UC-Net~\cite{zhang2020uc}, JL-DCF\cite{fu2020jl}, CoNet~\cite{ji2020accurate}, DANet~\cite{zhao2020single}, EBFSP\cite{huang2021employing},CDNet\cite{jin2021cdnet},
HAINet\cite{li2021hierarchical}, RD3D\cite{chen2021rd3d}, DSA2F\cite{Sun2021DeepRS},
MMNet\cite{gao2021unified} and
VST\cite{liu2021visual}. To ensure the fairness of the comparison results, the saliency maps of the evaluation are provided by the authors or generated by running source codes.

\begin{figure*}[!htp]
\centering
\begin{tabular}{ccc}
\includegraphics[width = 0.33\textwidth]{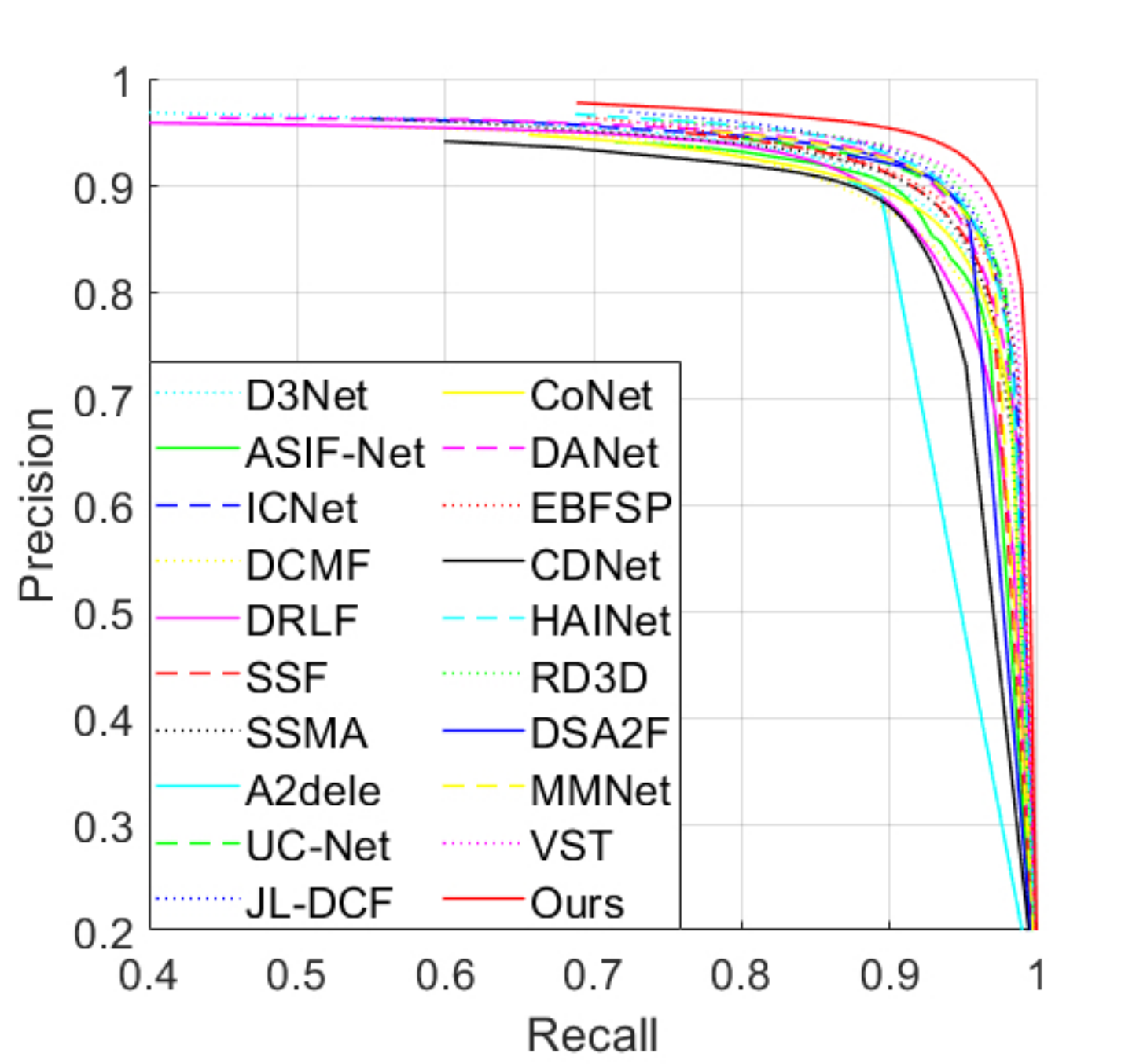}&\includegraphics[width = 0.33\textwidth]{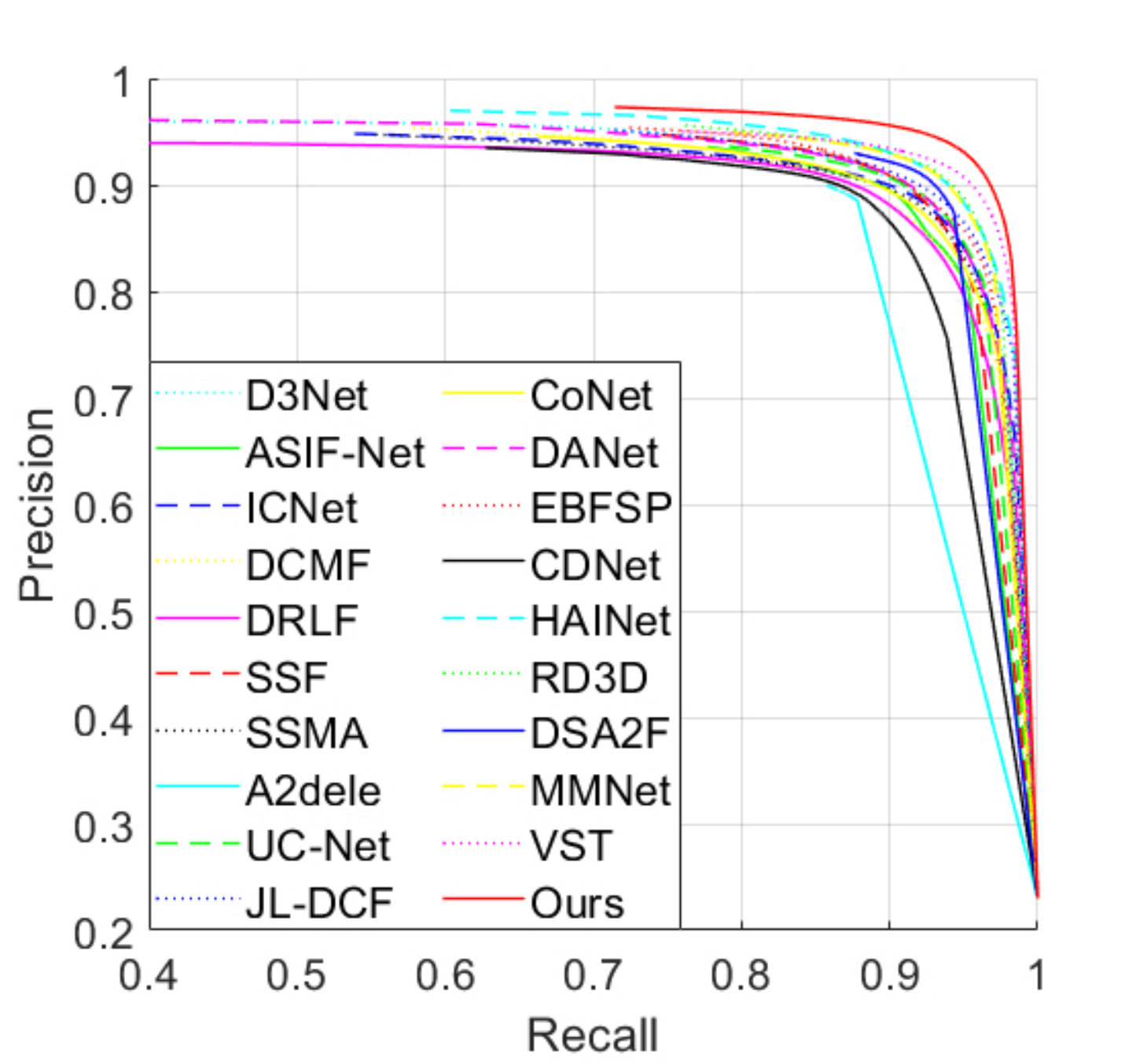}&\includegraphics[width = 0.33\textwidth]{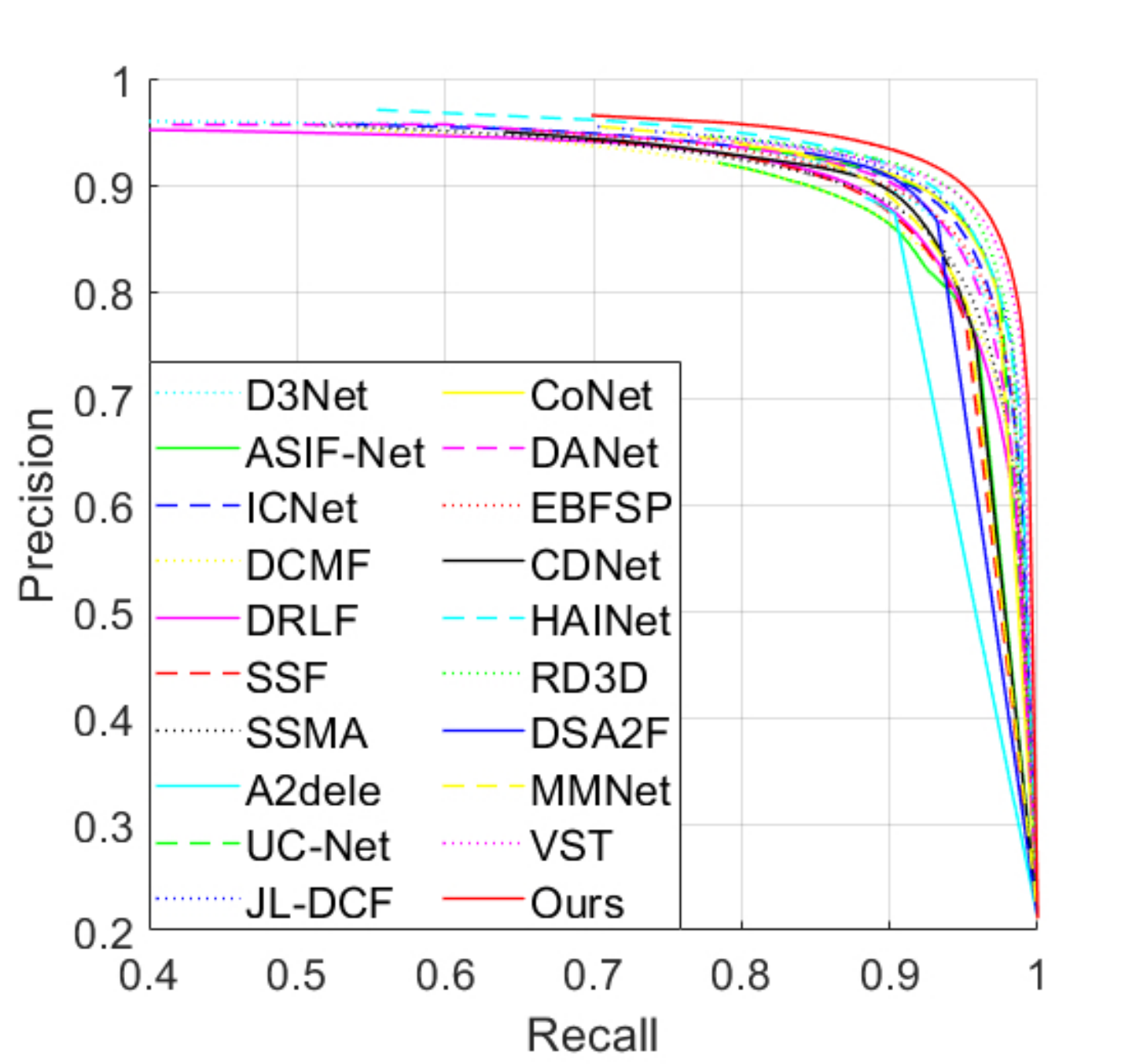}\\
(a) NLPR dataset&(b) NJU2K dataset&(c) STERE dataset\\
\end{tabular}
\begin{tabular}{ccc}
\includegraphics[width =
0.33\textwidth]{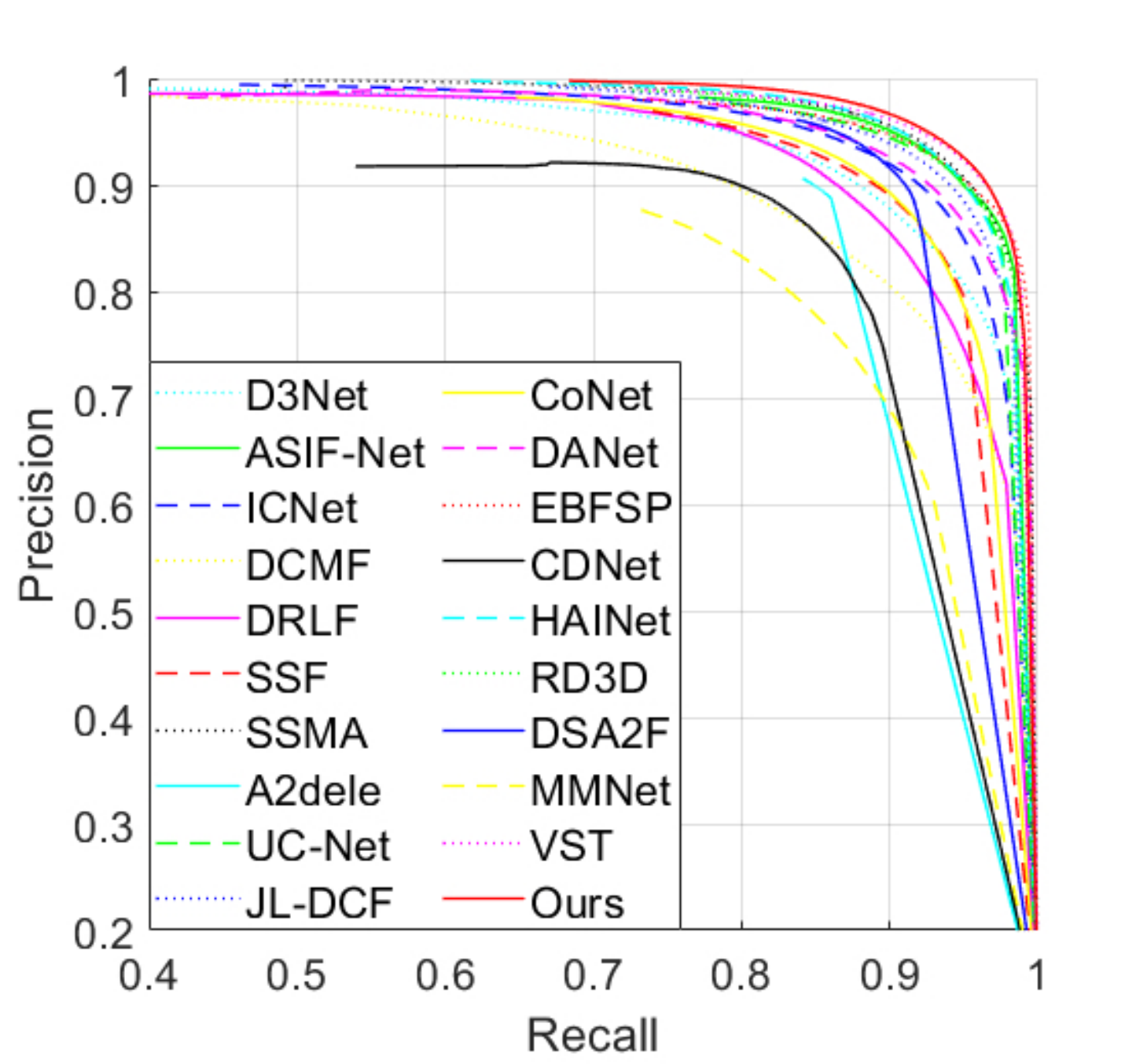}&\includegraphics[width = 0.33\textwidth]{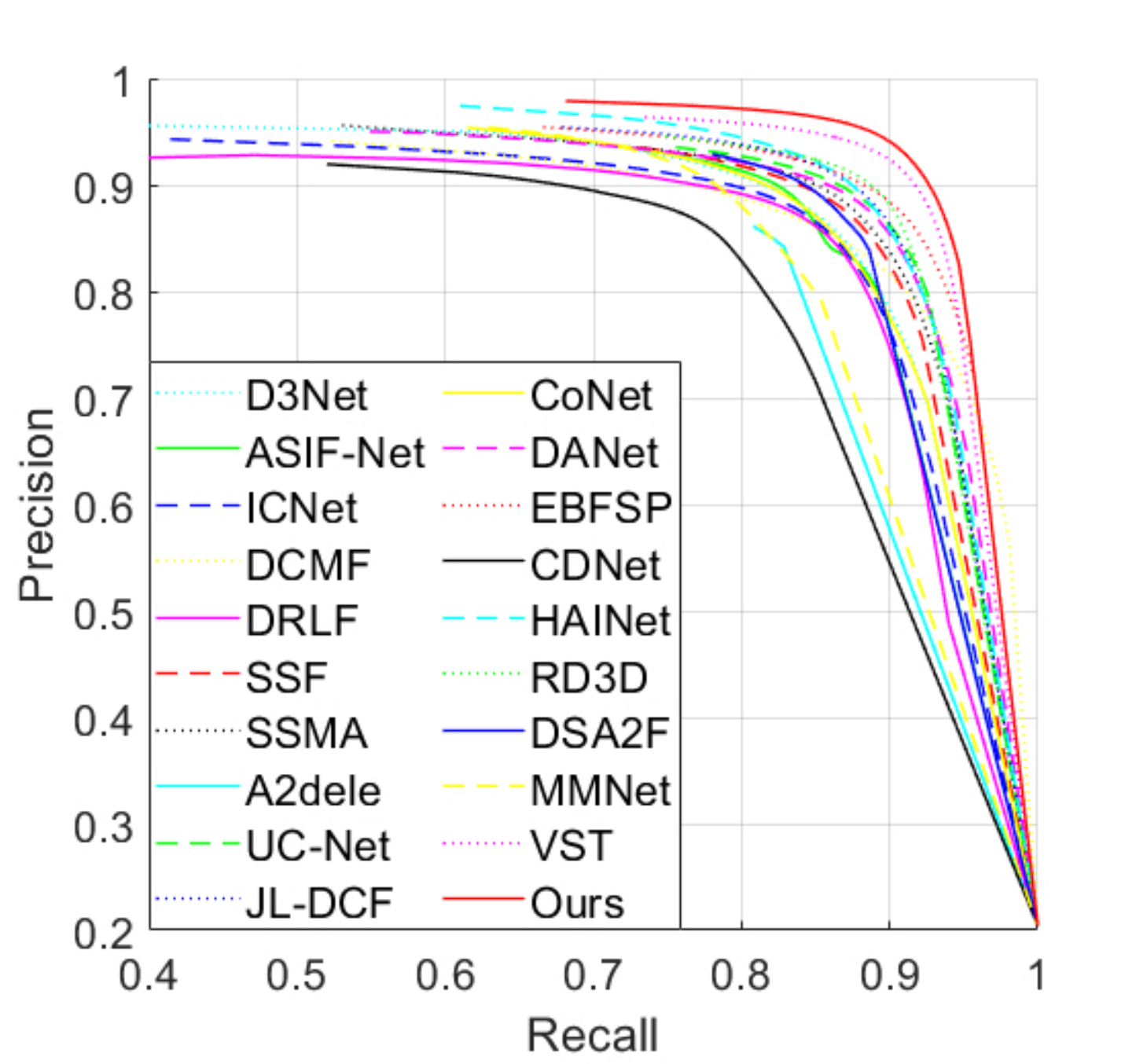}&\includegraphics[width = 0.33\textwidth]{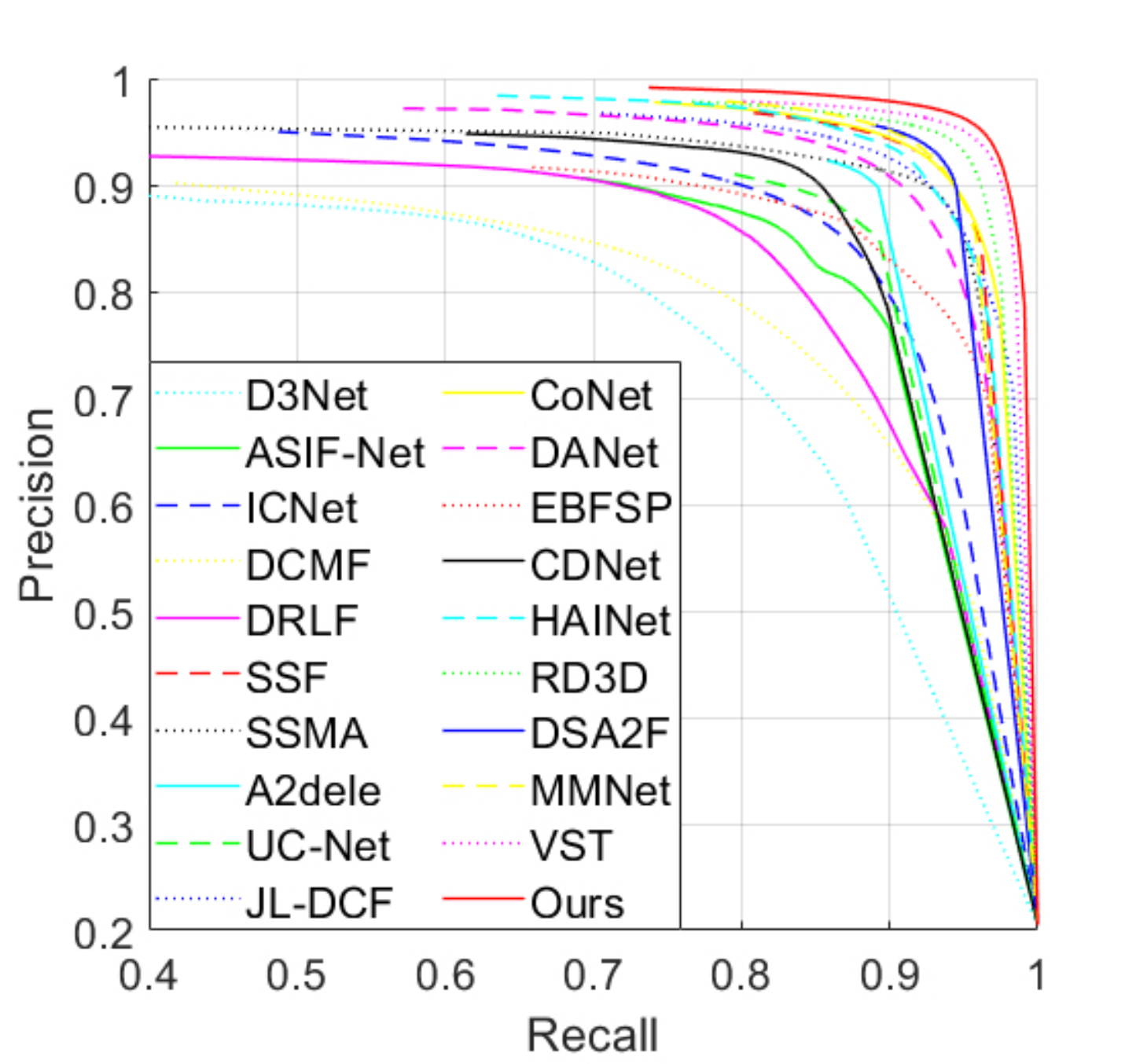}\\
(d) DES dataset &(e) SIP dataset&(f) DUT dataset\\
\end{tabular}
\caption{P-R curves comparison of different models on six RGB-D datasets. Our SwinNet represented by red solid line outperforms SOTA models.}
\label{fig:RGBDPRComparison}
\end{figure*}

\textbf{Quantitative Evaluation.} Fig.\ref{fig:RGBDPRComparison} shows the comparison results on PR curve. Table.\ref{tab:RGBDcomparison} shows the quantitative comparison results of four
evaluation metrics.
As can be clearly observed from figure that our curves are significant better than the others on NLPR, NJU2K, STERE, SIP and DUT datasets, slightly on DES dataset. It benefits from the choose of backbone, spatial alignment and channel re-calibration of two modalities and edge guidance. Meanwhile, the table also gives the consistent results. The performance is improved with a large margin on NLPR, NJU2K, STERE, SIP and DUT datasets, and has a little effectiveness on DES dataset.
Compared with  transformer-based method VST\cite{liu2021visual}, S-measure, F-measure, E-measure and MAE are improved about 0.007, 0.017, 0.010 and 0.005 on average. The PR curve and evaluation metrics all verify the effectiveness and advantages of our proposed method in RGB-D SOD task.

\begin{figure*}[!htp]
	\centering \includegraphics[width=1\textwidth]{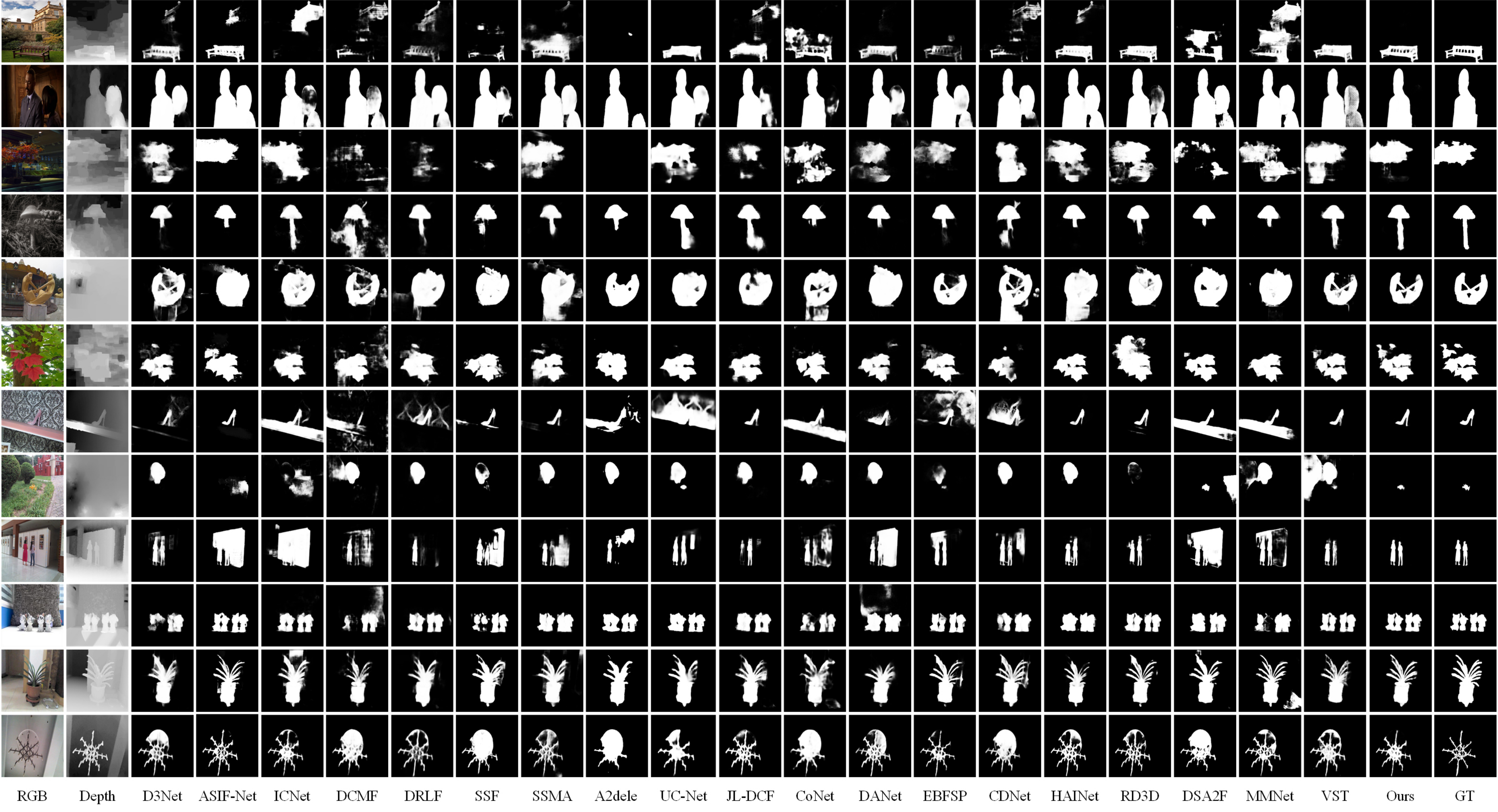}
	\caption{Visual comparison with SOTA RGB-D models. Our SwinNet is outstanding in some challenging cases: similar foreground and background (1$^{st}$-2$^{nd}$rows), complex scene (3$^{rd}$-4$^{th}$rows), depth image with low quality (5$^{th}$-6$^{th}$rows), small
object (7$^{th}$-8$^{th}$rows), multiple objects (9$^{th}$-10$^{th}$rows) and fine-grained object ($11^{th}_{}$-$12^{th}_{}$rows).}
\label{fig:RGBDVisual_compare}
\end{figure*}

\begin{table*}[!htp]
  \centering
  \fontsize{7}{10}\selectfont
  \renewcommand{\arraystretch}{1}
  \renewcommand{\tabcolsep}{0.2mm}
  \caption{S-measure, adaptive F-measure, adaptive E-measure, MAE comparisons with different RGB-D models.  The best result is in bold.}
\label{tab:RGBDcomparison}
  \begin{tabular}{c|c|ccccccccccccccccccccc}
  \hline\toprule
    \multirow{2}{*}{Datasets} & \multirow{2}{*}{Metric}
    & D3Net &ASIF-Net & ICNet & DCMF & DRLF & SSF & SSMA & A2dele & UC-Net & JL-DCF & CoNet & DANet  & EBFSP & CDNet & HAINet & RD3D & DSA2F & MMNet &VST & SwinNet\\
    &  & TNNLS20 & {TCYB20} & TIP20 & TIP20 & TIP20 & CVPR20 & CVPR20 & CVPR20 & CVPR20  & CVPR20 & ECCV20 & ECCV20 & TMM21 & TIP21 & TIP21 & AAAI21 & CVPR21  & {TCSVT21} & arXiv & Ours \\
    \midrule

\multirow{4}{*}{NLPR}
    & S$\uparrow$         & .912 &.909& .923 & .900 & .903 & .914 & .915 & .896 & .920 &.925 & .908 & .920  & .915 & .902 & .924 & .930 & .918 &.925 & .931 &  \textbf{.941}\\

    & F$_\beta \uparrow$  & .861 & {.869} & .870 & .839 & .843 & .875 & .853 & .878 & .890 & .878 & .846 & .875  & .897 & .848 & .897 & .892 & .892 & {.889} & .886 &  \textbf{.908}\\

    & $E_{\xi}\uparrow$   & .944  & {.944} & .944 & .933 & .936 & .949 & .938 & .945 & .953 & .953 & .934 & .951  & .952 & .935 & .957 & .958 & .950 & { .950} & .954 &  \textbf{.967}\\

    & MAE$\downarrow$     & .030  & {.029} & .028 & .035 & .032 & .026 & .030 & .028 & .025 & .022 & .031 & .027  & .026 & .032 & .024 & .022 & .024 & {.024} & .023 &  \textbf{.018}\\
    \midrule

\multirow{4}{*}{NJU2K}
    & S$\uparrow$         & .901  & { .891} & .894 & .889 & .886 & .899 & .894 & .869 & .897 & .902 & .895 & .899  & .903 & .885 & .912 & .916 & .904 & {.911} & .922 &\textbf{.935}\\

    & F$_\beta \uparrow$  & .865  & {.877} & .868 & .859 & .849 & .886 & .865 & .874 & .889 & .885 & .872 & .871  & .894 & .866 & .900 & .901 & .898 & {.900} & .899 &  \textbf{.922}\\

    & $E_{\xi}\uparrow$   & .914 & {.907}  & .905 & .897 & .901 & .913 & .896 & .897 & .903 & .913 & .912 & .908  & .907 & .911 & .922 & .918 & .922 & {.919} & .914 &  \textbf{.934}\\

    & MAE$\downarrow$     & .046 & {.047} & .052 & .052 & .055 & .043 & .053 & .051 & .043 & .041 & .046 & .045  & .039 & .048 & .038 & .036 & .039 & {.038} & .034 &\textbf{.027} \\
    \midrule

\multirow{4}{*}{STERE}
    & S$\uparrow$         & .899 & {.874}  & .903 & .883 & .888 & .887 & .890 & .878 & .903 & .903 & .905 & .901  & .900 & .896 & .907 & .911 & .897 & {.891} & .913 & \textbf{.919}\\

    & F$_\beta \uparrow$  & .859 & {.852} & .865 & .841 & .845 & .867 & .855 & .874 & .885 & .869 & .884 & .868  & .870 & .873 & .885 & .886 & .893  & {.880} & .878 & \textbf{.893}\\

    & $E_{\xi}\uparrow$   & .920 & {.908} & .915 & .904 & .915 & .921 & .907 & .915 & .922 & .919 & .927 & .921  & .912 & .922 & .925 & .927 & .927 & { .924} & .917 &  \textbf{.929}\\

    & MAE$\downarrow$     & .046 & {.051} & .045 & .054 & .050 & .046 & .051 & .044 & .039  & .040 & .037 & .043  & .045 & .042 & .040 & .037 & .039 & {.045}  & .038 &  \textbf{.033}\\
    \midrule

\multirow{4}{*}{DES}
    & S$\uparrow$         & .898  & {.934}& .920 & .877 & .895 & .905 & .941 & .885 & .933  & .931 & .911 & .924  & .937 & .875 & .935 & .935 & .916 & {.830} & .943    & \textbf{.945}\\

    & F$_\beta \uparrow$  & .870  & {.915} & .889 & .820 & .868 & .876 & .906 & .865 & .917 & .900 & .861 & .899  & .913 & .839 & .924 & .917 & .901 &{.746} & .917    &  \textbf{.926}\\

    & $E_{\xi}\uparrow$   & .951 & {.974} & .959 & .923 & .954 & .948 & .974 & .922 & .974 & .969 & .945 & .968  & .974 & .921 & .974 & .975 & .955  &{.893} & .979    & \textbf{.980}\\

    & MAE$\downarrow$     & .031 &{.019} & .027 & .040 & .030 & .025 & .021 & .028 & .018 & .020 & .027 & .023  & .018 & .034 & .018 & .019 & .023 & {.058}  & .017    & \textbf{.016}\\
    \midrule

\multirow{4}{*}{SIP}
    & S$\uparrow$         & .860 & {.857} & .854 & .859 & .850 & .868 & .872 & .826 & .875 & .880 & .858 & .875  & .885 & .823 & .880 & .885 & .862 & {.836} & .904 &\textbf{.911}\\

    & F$_\beta \uparrow$  & .835 & {.847} & .836 & .819 & .813 & .851 & .854 & .825 & .868  & .873 & .842 & .855  & .869 & .805 & .875 & .874 & .865 &{.839} & .895 &\textbf{.912}  \\

    & $E_{\xi}\uparrow$   & .902  & { .895} & .899 & .898 & .891 & .911 & .911 & .892 & .913 & .921 & .909 & .914  & .917 & .880 & .919 & .920 & .908  &{.882} & .937 &\textbf{.943} \\
    & MAE$\downarrow$     & .063 & {.061} & .069 & .068 & .071 & .056 & .057 & .070 & .051 & .049  & .063 & .054  & .049 & .076 & .053 & .048 & .057 &{.075} & .040 & \textbf{.035} \\
    \bottomrule
    \hline
\end{tabular}
\end{table*}
\textbf{Qualitative Evaluation.} To make the qualitative comparisons, we show some visual examples in Fig.\ref{fig:RGBDVisual_compare}.
It can be observed that our method has the better detection results than other methods in some challenging cases: similar foreground and background ($1^{st}_{}$-$2^{nd}_{}$ rows), complex scene ($3^{rd}_{}$-$4^{th}_{}$ rows), depth image with low quality ($5^{th}_{}$-$6^{th}_{}$ rows), small object ($7^{th}_{}$-$8^{th}_{}$ rows) and multiple objects ($9^{th}_{}$-$10^{th}_{}$ rows). %
In addition, our approach can produce more fine-grained details as highlighted in the salient region ($11^{th}_{}$-$12^{th}_{}$ rows). 
The visual examples indicate that our approach can better locate salient objects and produce more accurate saliency maps.

\subsubsection{RGB-T SOD}
For RGB-T SOD,
our model is compared with some SOTA RGB-T SOD algorithms, including
MTMR\cite{wang2018rgb}, 
M3S-NIR\cite{tu2019m3s}, 
SGDL\cite{tu2019rgb}, 
ADF\cite{tu2020rgbt}, 
ECFFNet\cite{zhou2021ecffnet}, 
MIDD\cite{tu2021multiinteractive},
MMNet\cite{gao2021unified},
CSRNet\cite{huo2021efficient},
CGFNet\cite{wang2021cgfnet}.
To ensure the fairness of the comparison results, the saliency maps of the evaluation are provided by the authors or generated by running source codes.

\begin{figure*}[!htp]
\centering
\begin{tabular}{ccc}
\includegraphics[width = 0.33\textwidth]{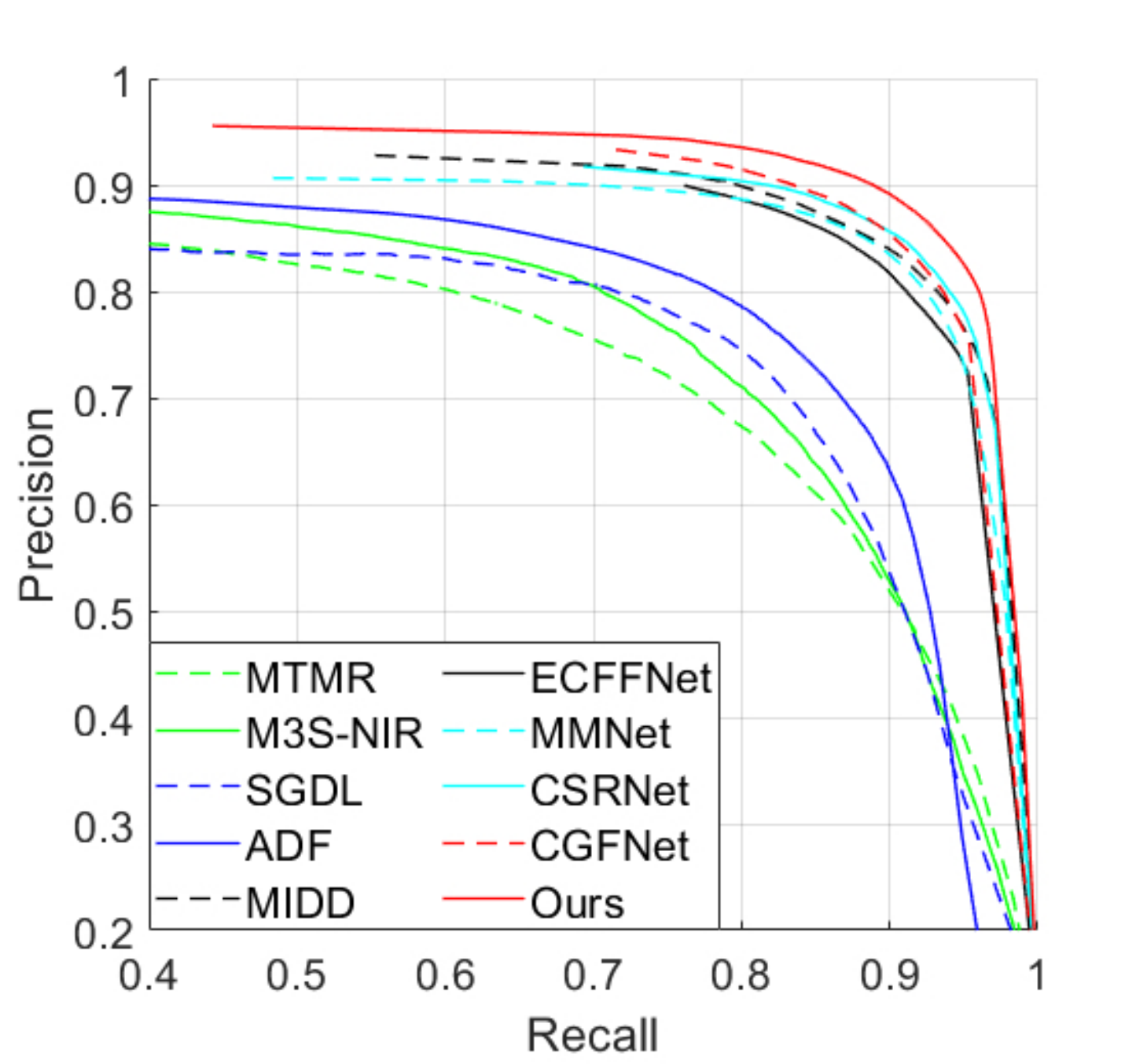}&\includegraphics[width = 0.33\textwidth]{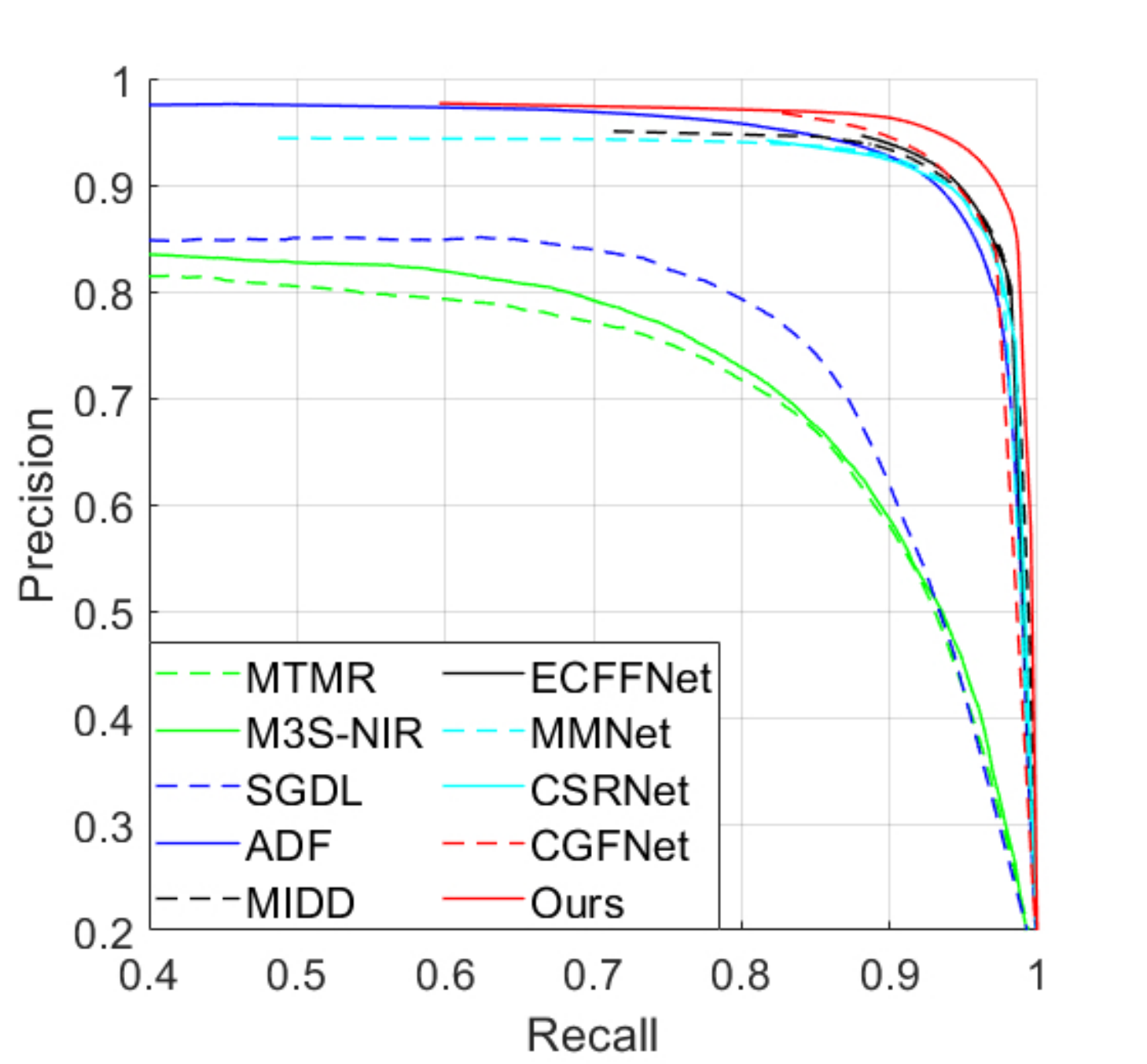}&\includegraphics[width = 0.33\textwidth]{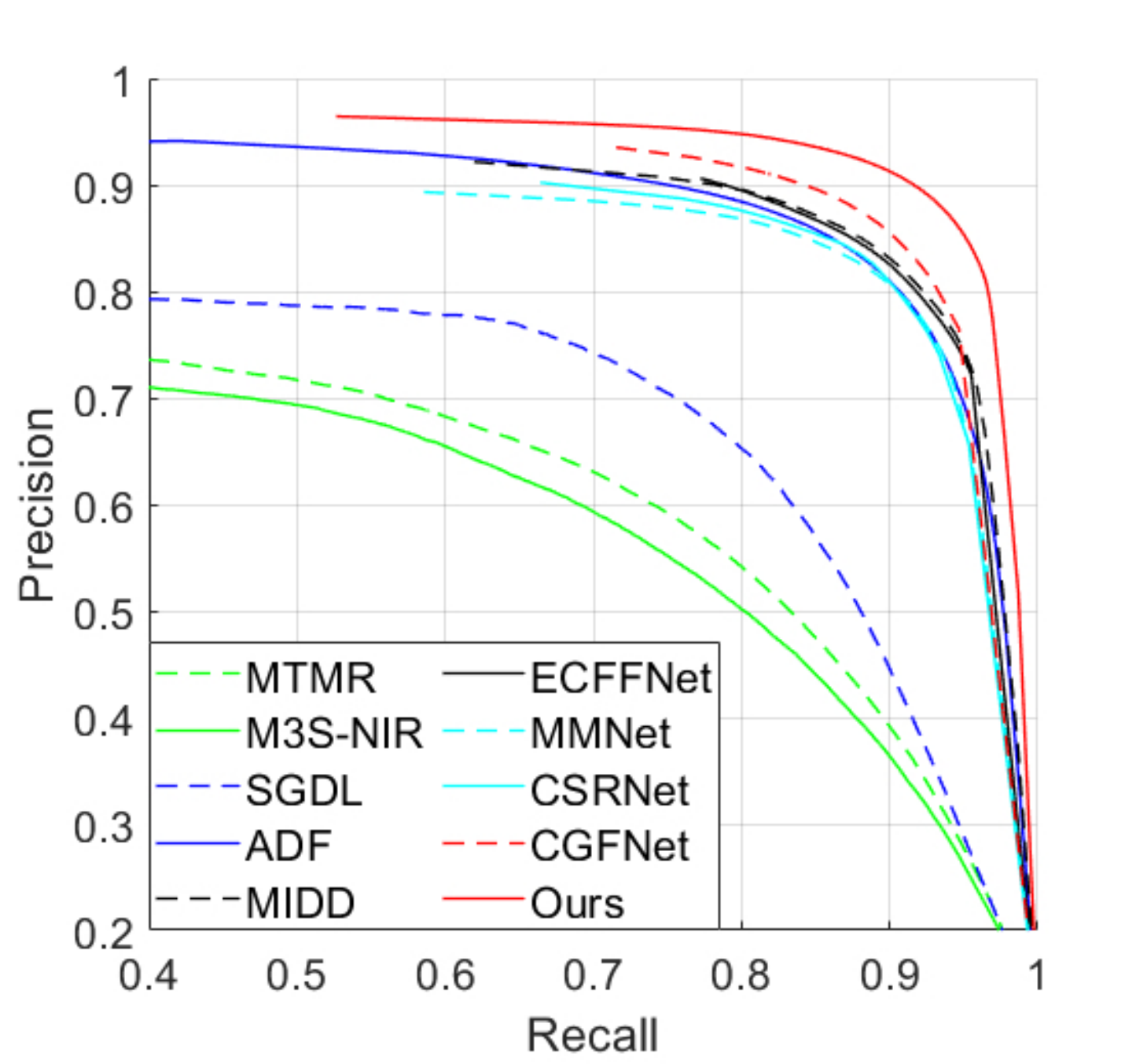}\\
(a) VT821 dataset&(b) VT1000 dataset&(c) VT5000 dataset\\
\end{tabular}
\caption{P-R curves comparison of different models on three RGB-T datasets. Our SwinNet represented by red solid line outperforms SOTA models.}
\label{fig:RGBTPRComparison}
\end{figure*}

\textbf{Quantitative Evaluation.} Fig.\ref{fig:RGBTPRComparison} shows the comparison results on PR curve. Table.\ref{tab:RGBTcomparison} shows the quantitative comparison results of four evaluation metrics.
As can be clearly found from figure that our curves are very high, which means that our method is superior to the others with a large margin. Furthermore, from the table, we can see that all the evaluation metrics are the best and our performance is significantly improved.
The PR curve and evaluation metrics all verify the effectiveness and advantages of our proposed method in RGB-T SOD task.

\begin{table*}[!htp]
  \centering
  \renewcommand{\arraystretch}{1}
  \renewcommand{\tabcolsep}{0.5mm}
  \caption{S-measure,  adaptive F-measure, adaptive E-measure,  MAE comparisons with different RGB-T models. The best result is in bold.}
\label{tab:RGBTcomparison}
  \begin{tabular}{c|c|cccccccccc}
  \hline\toprule
Datasets & Metric & MTMR  & M3S-NIR  & SGDL & ADF & MIDD &  ECFFNet & {MMNet}  & {CSRNet} & {CGFNet} & SwinNet \\
    &  & IGTA18 & MIPR19 & TMM19 & Arxiv20  & TIP21 & TCSVT21  &{TCSVT21} &{TCSVT21}& {TCSVT21}& Ours \\
        \midrule
\multirow{5}{*}{VT821}
    & S$_\alpha \uparrow$    & .725 & .723 & .765 & .810 & .871 & .877 & {.875} & {.885} & {.881}
    & \textbf{.904}\\
    & F$_\beta \uparrow$     & .662 & .734 & .730 & .716 & .804 & .810 &  {.798} & .830 & {.845} &\textbf{.847} \\
    & E$_\epsilon \uparrow$  & .815 & .859 & .847 & .842 & .895 & .902 & {.893} & {.908} & {.912}
    & \textbf{.926} \\
    & MAE$\downarrow$        & .108 & .140 & .085 & .077 & .045& .034  & {.040} & {.038} & {.038}
    & \textbf{.030} \\
    \midrule
\multirow{5}{*}{VT1000}
    & S$_\alpha \uparrow$    & .706 & .726 & .787 & .910 & .915 &  .923 & {.917} & {.918} & {.923} &\textbf{.938} \\
    & F$_\beta \uparrow$     & .715 & .717 & .764 & .847 & .882 & .876  & {.863} & {.877} & \textbf{{.906} }& .896\\
    & E$_\epsilon \uparrow$  & .836 & .827 & .856 & .921 &  .933 & .930  & {.924} & {.925} & {.944} &\textbf{.947}\\
    & MAE$\downarrow$        & .119 & .145 & .090 & .034 &.027 &  .021   & {.027} &{.024} & {.023} &\textbf{.018}\\
    \midrule
\multirow{5}{*}{VT5000}
    & S$_\alpha \uparrow$    & .680 & .652 & .750 & .863 & .867 & .874  & {.864} & {.868} & {.883}
    & \textbf{.912}\\
    & F$_\beta \uparrow$     & .595 & .575 & .672 & .778 &.801 & .806   & {.785} & {.810} & {.851}
    & \textbf{.865}\\
    & E$_\epsilon \uparrow$  & .795 & .780 & .824 & .891 & .897 & .906  & {.890} & {.905} & {.922}
    & \textbf{.942}\\
    & MAE$\downarrow$ & .114 & .168 & .089 & .048 & .043 & .038 & {.043} & {.042} & {.035}
    & \textbf{.026} \\
    \bottomrule\hline
  \end{tabular}
\end{table*}

\begin{figure*}[!htp]
	\centering	\includegraphics[width=0.7\textwidth]{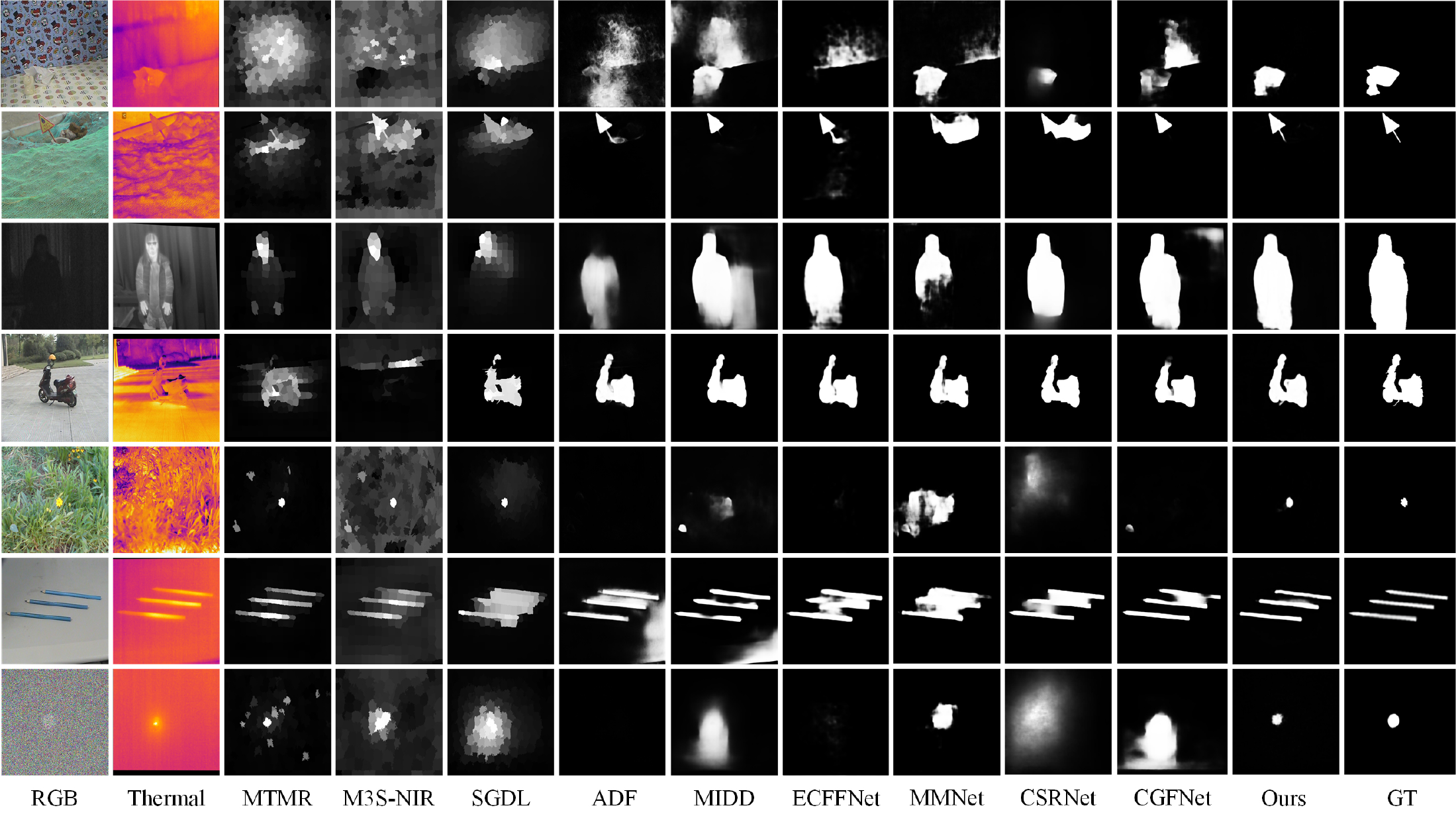}
	\caption{Visual comparison with SOTA RGB-T models. Our SwinNet is outstanding in some challenging cases: similar foreground and background ($1^{st}$row), complex scene ($2^{nd}$row), poor illuminance ($3^{rd}$row), low contrast of thermal image ($4^{th}$row),
small object ($5^{th}$row), multiple objects ($6^{th}$row) and noise disturbance object ($7^{th}$row).}
\label{fig:RGBTvisual_compare}
\end{figure*}

\textbf{Qualitative Evaluation.} To make the qualitative comparisons, we show some visual examples in Fig.\ref{fig:RGBTvisual_compare}.
It can be observed that our method has the better detection results than other methods in some challenging cases: similar foreground and background ($1^{st}$ row), complex scene ($2^{nd}$ row), poor illuminance ($3^{rd}$ row), low contrast of thermal image ($4^{th}$ row),
small object ($5^{th}$ row) and multiple objects ($6^{th}$ row).
In addition, our approach is robust to noise disturbance, which can be seen in the $7^{th}$ row.
These all indicate that our approach can better adapt to different scenes, and work well by cross-modality fusion.

\subsection{Ablation studies}
We conduct ablation studies on RGB-D SOD to verify all of components.

\subsubsection{The effectiveness of Swin Transformer backbone}

\begin{figure*}[!htp]
	\centering
\includegraphics[width=0.7\textwidth]{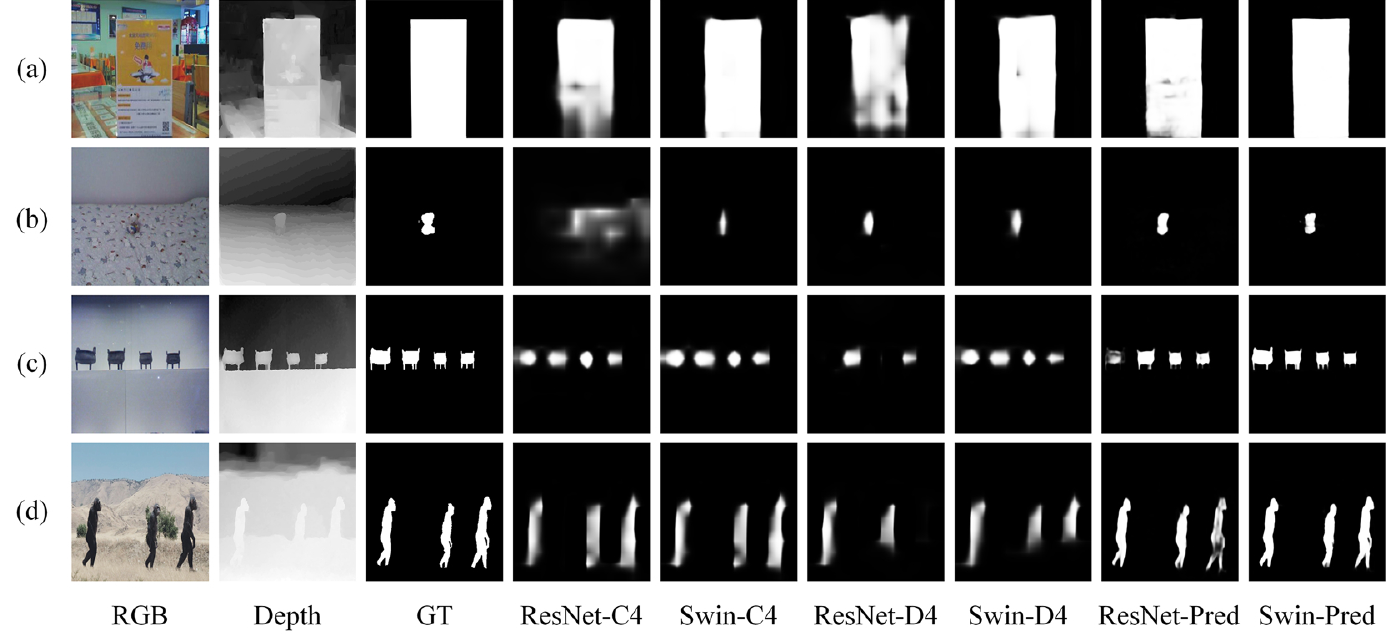}
	\caption{Visual comparison between ResNet and Swin Transformer. From the left to the right, there are RGB image, depth image, ground truth (GT), the color feature in the fourth layer of ResNet (ResNet-C4), the color feature in the fourth layer of Swin Transformer (Swin-C4), the depth feature in the fourth layer of ResNet (ResNet-D4), the depth feature in the fourth layer of Swin Transformer (Swin-D4), the prediction saliency map of ResNet (ResNet-Pred), the prediction saliency map of Swin Transformer (Swin-Pred).}  \label{fig:Compare_ResNet_Swin}
\end{figure*}

We replaces Swin Transformer backbone with some CNN backbones (e.g., ResNet-50\cite{he2016deep}, Res2Net-50\cite{gao2019res2net}, ResNet-101\cite{he2016deep}, Res50+ViT16\cite{chen2021transunet}) and transformer backbones (e.g., T2T-14\cite{yuan2021tokens} and PVT-M\cite{wang2021pyramid}) to check the effectiveness of backbones.
From Table.~\ref{tab:BackboneAblation}, we can find that the use of Swin Transformer significantly improves the detection performance. It profits from the integration of locality merit of CNN and global-aware ability of transformer.
We also give some visual comparison of ResNet101 and Swin Transformer in Fig.~\ref{fig:Compare_ResNet_Swin}. From the left to the right, there are RGB image, depth image, ground truth (GT), the color feature in the fourth layer of ResNet (ResNet-C4), the color feature in the fourth layer of Swin Transformer (Swin-C4), the depth feature in the fourth layer of ResNet (ResNet-D4), the depth feature in the fourth layer of Swin Transformer (Swin-D4), the prediction saliency map of ResNet (ResNet-Pred), the prediction saliency map of Swin Transformer (Swin-Pred). From the Fig.~\ref{fig:Compare_ResNet_Swin} (a)(b), we can discover that ResNet-C4 is interior to Swin-C4, so as to generate the blurry prediction saliency map.
From the Fig.~\ref{fig:Compare_ResNet_Swin} (c)(d), we can find that some small objects are ignored in ResNet-D4. It may be caused by the larger receptive field in convolution neural network. Equipped with the long-range dependency merit, Swin-D4 shows  salient features with more integrity. Certainly, Swin-Pred shows the better result than ResNet-Pred.

\begin{table*}[!htp]
\caption{Effectiveness analysis of backbone network, including ResNet-50, Res2Net-50, ResNet-101, ResNet-50+ViT16, T2T-14, PVT-M and Swin-B. The best result is in bold.}

  \centering
  \fontsize{8}{10}\selectfont
  \renewcommand{\arraystretch}{1}
  \renewcommand{\tabcolsep}{0.8mm}
\begin{tabular}{c|cccc|cccc|cccc|cccc}
\hline\toprule
   \multirow{2}{*}{\centering Backbone} & 
   \multicolumn{4}{c|}{\centering NLPR} & \multicolumn{4}{c|}{\centering NJU2K} & \multicolumn{4}{c|}{\centering STERE} & \multicolumn{4}{c}{\centering SIP}\\
     & S$\uparrow$
     & F$_\beta$ $\uparrow$ &$E_{\xi}\uparrow$
     & MAE$\downarrow$ & S$\uparrow$
     & F$_\beta$ $\uparrow$ &$E_{\xi}\uparrow$
     & MAE$\downarrow$ & S$\uparrow$
     & F$_\beta$$\uparrow$&$E_{\xi}\uparrow$ & MAE$\downarrow$
     & S$\uparrow$
     & F$_\beta$$\uparrow$ &$E_{\xi}\uparrow$& MAE$\downarrow$ \\
    \hline
    ResNet-50
    &.925&.878&.948&.026&.911 &.894&.916&.039 &	.896 &.872&.920 &.044&.892&.879&.926&.046\\

      Res2Net-50 
    &.905 &.813& .924 &.036&.897
     &.840& .891&.054 &.880
      &.816& .892 &.061 &.870
     &.840& .905 &.065 \\

ResNet-101  &.924
    &.884& .955 &.024&.920
     &.904& .922&.034 &.885
      &.861& .918 &.049 &.897
     &.888& .931 &.043 \\

      Res-50+ViT16
 &.932 &.892& .960 &.021&.922
     &.904& .918&.033 &.903
      &.869& .917 &.041 &.894
     &.891& .930 &.046 \\

     T2T-14
&.928
    &.880& .958 &.022&.915
     &.893& .919&.037 &.894
      &.856& .918 &.044 &.897
     &.887& .931 &.045 \\

     PVT-M
 &.925
    &.879& .956 &.023&.917
     &.898& .921&.036 &.901
      &.869& .922 &.042 &.893
     &.888& .932 &.043 \\

      Swin-B 
 &\textbf{.941}
    &\textbf{.908} & \textbf{.967} &\textbf{.018} &\textbf{.935}
    &\textbf{.922}& \textbf{.934} &\textbf{.027} &\textbf{.919}
    &\textbf{.893}& \textbf{.929} &\textbf{.033} &\textbf{.911}
    &\textbf{.912}& \textbf{.943} &\textbf{.035}
    \\
    \bottomrule
    \hline
\end{tabular}
\label{tab:BackboneAblation}
\end{table*}

\subsubsection{The effectiveness of spatial alignment and channel re-calibration module}
\begin{figure}[!htp]
	\centering	\includegraphics[width=1\linewidth]{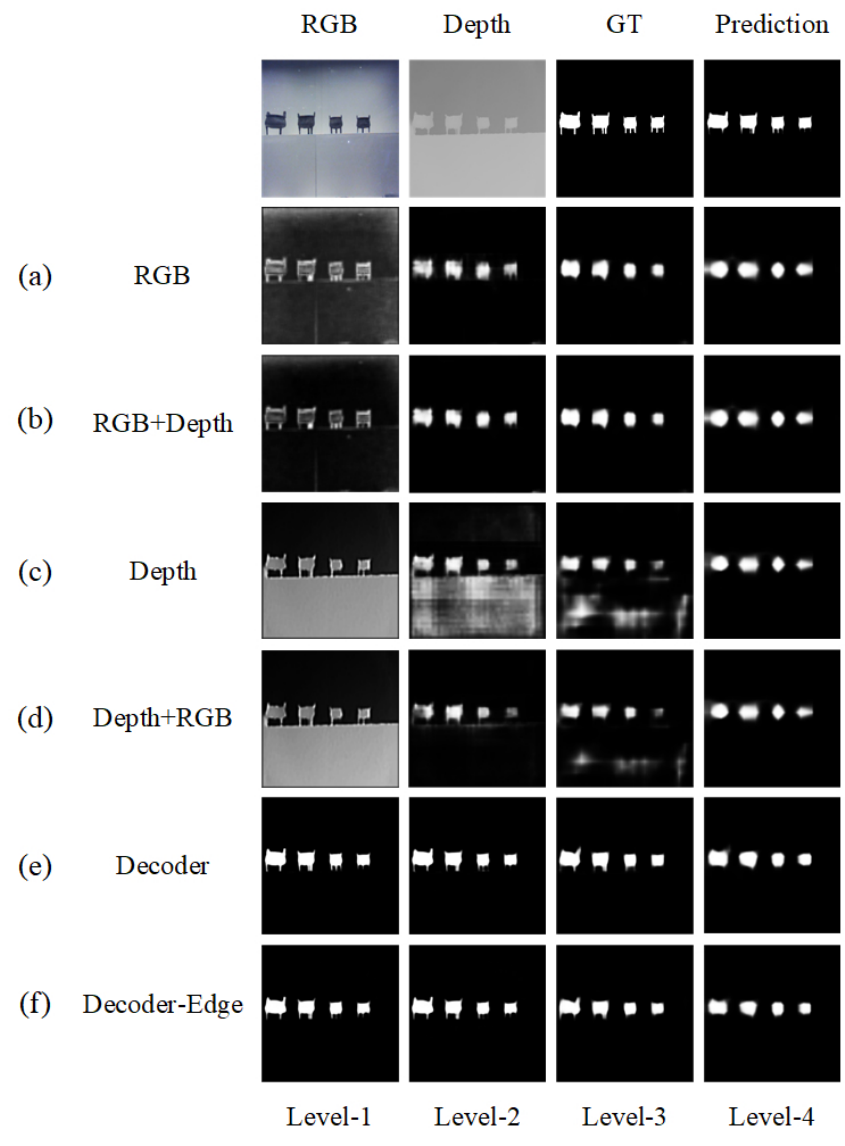}
	\caption{Visual comparison about the effectiveness of spatial alignment and channel re-calibration module and edge guidance. }  \label{fig:ablationVisual}
\end{figure}

Fig.~\ref{fig:ablationVisual} shows visual comparison of some ablation studies. From left to right, there are RGB image, depth image, ground truth saliency map, prediction saliency map in the first line. In  other lines, there are features in different layers, corresponding with the color features from backbones $\{ST_i^c\}_{i=1}^4$,  the color features after spatial alignment and channel re-calibration module $\{F_i^c\}_{i=1}^4$, the depth features from backbones $\{ST_i^d\}_{i=1}^4$, the depth features after spatial alignment and channel re-calibration module $\{F_i^d\}_{i=1}^4$, the features from decoder $\{FF_i\}_{i=1}^4$. Last, a group of features in the decoder without edge guidance are shown in the last line.

From the comparison between Fig.~\ref{fig:ablationVisual} (a) and (b), we find that the color features after spatial alignment and channel re-calibration are purified and  the noises are obviously reduced, especially in the first column.
From the comparison between Fig.~\ref{fig:ablationVisual} (c) and (d), we find that the depth features with the help of color features are close to ground truth, and salient region are misjudged less, especially in the second and third column.

Furthermore, we replace spatial alignment and channel re-calibration module with Depth-enhanced Module (DEM) in BBS-Net~\cite{fan2020bbs} which consists of the similar channel attention and spatial attention but no alignment operation to verify  the effectiveness of spatial alignment and channel re-calibration module.
From Table.~\ref{tab:SABRAblation}, we can see that
our S-measure, F-measure and E-measure and MAE wins about 0.006, 0.006, 0.005 and 0.002 when compared with DEM. The proposed spatial alignment and channel re-calibration module enhances the feature representation of color and depth images by the intra-layer interaction between two modalities and attentional weight assignment.

\begin{table*}[!htp]
\caption{Effectiveness analysis of spatial alignment and channel re-calibration module. DEM denotes the model with Depth-enhanced Module in BBS-Net~\cite{fan2020bbs} instead of our spatial alignment and channel re-calibration module. The best result is in bold.}
\fontsize{8}{10}\selectfont
  \renewcommand{\arraystretch}{1}
  \renewcommand{\tabcolsep}{0.8mm}
\begin{center}
\begin{tabular}{c|cccc|cccc|cccc|cccc}
\hline\toprule
   \multirow{2}{*}{\centering Variant} & \multicolumn{4}{c|}{\centering NLPR} & \multicolumn{4}{c|}{\centering NJU2K} & \multicolumn{4}{c|}{\centering STERE} & \multicolumn{4}{c}{\centering SIP}\\
     & S$\uparrow$
     & F$_\beta$ $\uparrow$ &$E_{\xi}\uparrow$
     & MAE$\downarrow$ & S$\uparrow$
     & F$_\beta$ $\uparrow$ &$E_{\xi}\uparrow$
     & MAE$\downarrow$ & S$\uparrow$
     & F$_\beta$$\uparrow$&$E_{\xi}\uparrow$ & MAE$\downarrow$
     & S$\uparrow$
     & F$_\beta$$\uparrow$ &$E_{\xi}\uparrow$& MAE$\downarrow$ \\
    \hline
    DEM  &.936
    &.903& .964 &.019&.930
     &.918& .929&.028 &.914
      &.888& .928 &.034 &.900
     &.901& .933 &.040 \\
    Ours  &\textbf{.941}
    &\textbf{.908} & \textbf{.967} &\textbf{.018} &\textbf{.935}
    &\textbf{.922}& \textbf{.934} &\textbf{.027} &\textbf{.919}
    &\textbf{.893}& \textbf{.929} &\textbf{.033} &\textbf{.911}
    &\textbf{.912}& \textbf{.943} &\textbf{.035}
    \\
    \bottomrule
    \hline
\end{tabular}
\end{center}
\label{tab:SABRAblation}
\end{table*}

\subsubsection{The effectiveness of edge guidance}
We remove the edge guidance in the decoder to verify its effectiveness.
From Fig.~\ref{fig:ablationVisual} (e) and (f), we can find the use of edge features enhances the detail of detected objects. Meanwhile, from Table.~\ref{tab:EdgeAblation}, we can also see that our S-measure, F-measure and E-measure and MAE are improved  about 0.004, 0.009, 0.006 and 0.003, respectively. It can further illustrate that edge guidance improves the performance of our proposed model to some extend.

\begin{table*}[!htp]
\caption{Effectiveness analysis of edge-guided decoder. The best result is in bold.}
  \begin{center}
\fontsize{8}{10}\selectfont
  \renewcommand{\arraystretch}{1}
  \renewcommand{\tabcolsep}{0.8mm}
\begin{tabular}{c|cccc|cccc|cccc|cccc}
\hline\toprule
   \multirow{2}{*}{\centering Variant} & \multicolumn{4}{c|}{\centering NLPR} & \multicolumn{4}{c|}{\centering NJU2K} & \multicolumn{4}{c|}{\centering STERE} & \multicolumn{4}{c}{\centering SIP}\\
     & S$\uparrow$
     & F$_\beta$ $\uparrow$ &$E_{\xi}\uparrow$
     & MAE$\downarrow$ & S$\uparrow$
     & F$_\beta$ $\uparrow$ &$E_{\xi}\uparrow$
     & MAE$\downarrow$ & S$\uparrow$
     & F$_\beta$$\uparrow$&$E_{\xi}\uparrow$ & MAE$\downarrow$
     & S$\uparrow$
     & F$_\beta$$\uparrow$ &$E_{\xi}\uparrow$& MAE$\downarrow$ \\
    \hline
    Without edge &.938
    &.901& .963 &.020&.928
     &.911& .922&.031 &\textbf{.919}
      &.887& .927 &.034 &.905
     &.900& .937 &.040 \\
    Ours  &\textbf{.941}
    &\textbf{.908} & \textbf{.967} &\textbf{.018} &\textbf{.935}
    &\textbf{.922}& \textbf{.934} &\textbf{.027} &\textbf{.919}
    &\textbf{.893}& \textbf{.929} &\textbf{.033} &\textbf{.911}
    &\textbf{.912}& \textbf{.943} &\textbf{.035}
    \\
    \bottomrule
    \hline
\end{tabular}
\end{center}
\label{tab:EdgeAblation}
\end{table*}

\subsubsection{The effectiveness of each modality}
To verify the contribution of each modality, we conduct the ablation study. From Table.\ref{tab:IndeMoComparisonRGBD} we can see that depth plays an obvious role in improving the performance from the first and third lines. Meanwhile, we also observe that depth information is interior to color cue in SOD performance when comparing the first and second lines. Especially, in STERE dataset, depth information plays a negative role because there are some depth images with low quality. The third line denoted as fusion result achieves the best results in a whole.
\begin{table*}[!htp]
  \centering
  \fontsize{8}{10}\selectfont
  \renewcommand{\arraystretch}{1}
  \renewcommand{\tabcolsep}{0.8mm}
  \caption{Ablation study about independent modality in RGB-D SOD. The best result is in bold. }
\label{tab:IndeMoComparisonRGBD}

\begin{tabular}{c|cccc|cccc|cccc|cccc}
   \hline\toprule
   \multirow{2}{*}{\centering Variant} & \multicolumn{4}{c|}{\centering NLPR} & \multicolumn{4}{c|}{\centering NJU2K} & \multicolumn{4}{c|}{\centering STERE} & \multicolumn{4}{c}{\centering SIP}\\
     & S$\uparrow$
     & F$_\beta$ $\uparrow$ &$E_{\xi}\uparrow$
     & MAE$\downarrow$ & S$\uparrow$
     & F$_\beta$ $\uparrow$ &$E_{\xi}\uparrow$
     & MAE$\downarrow$ & S$\uparrow$
     & F$_\beta$$\uparrow$&$E_{\xi}\uparrow$ & MAE$\downarrow$
     & S$\uparrow$
     & F$_\beta$$\uparrow$ &$E_{\xi}\uparrow$& MAE$\downarrow$ \\
    \hline
    RGB  &.932 &.869 &.955 &.024 &.921 &.901 &.919 &.036 &\textbf{.923} &\textbf{.898} &.926 &.034
    &.902 &.890 &.933 &.042\\

    Depth &.896 &.837 &.937 &.034 &.888 &.862 &.899 &.051 &.768 &.744 &.855 &.093 &.884 &.881 &.923 &.050 \\

     RGB+Depth &\textbf{.941} &\textbf{.908} &\textbf{.967} &\textbf{.018} &\textbf{.935} &\textbf{.922} &\textbf{.934} &\textbf{.027} &.919 &.893 &\textbf{.929 }&\textbf{.033} &\textbf{.911} &\textbf{.912} &\textbf{.943} &\textbf{.035}\\

    \bottomrule
    \hline
\end{tabular}

\end{table*}

\subsubsection{Model complexity analysis}
Model size of SwinNet is 198.7M parameters. Its computation cost is about 124.3G FLOPs and inference speed is about 10 FPS including all the  IO and preprocessing.
Its complexity is high.
From Table.\ref{tab:SwinNetModuleCostAblation}, we can find the computation cost mainly exists in Swin Transformer backbone. SwinNet-fuse denotes SwinNet removing spatial alignment and channel re-calibration module, SwinNet-edge denotes SwinNet removing edge-aware module, SwinNet-decoder denotes SwinNet removing edge-guided decoder.
Spatial alignment and channel re-calibration module and edge-aware module nearly spend no computation cost. Edge-guided decoder cost a little computation due to some convolution operations during upsampling process.
The majority of cost exists in two Swin Transformer backbones.

 \begin{table}[!htp]
\caption{Ablation study about model size and computation cost. }
  \centering
  \fontsize{8}{10}\selectfont
  \renewcommand{\arraystretch}{1}
  \renewcommand{\tabcolsep}{0.2mm}
\begin{tabular}{c|cccccccccccccc}
  \hline\toprule
    Methods
    & SwinNet & SwinNet-fuse & SwinNet-edge & SwinNet-decoder 
     \\
    \midrule
\multirow{1}{*}{Params(M)}
     &198.7  &198.3  &198.4  &173.6 
      \\
    \midrule

\multirow{1}{*}{FLOPs(G)}
    &124.3  &124.3  &122.4  &88.9 
    \\

    \bottomrule
    \hline
\end{tabular}
\label{tab:SwinNetModuleCostAblation}
\end{table}

\section{Conclusions}
Inspired by the success of transformer, it is introduced to drive RGB-D and RGB-T SOD. SwinNet  achieves SOTA performance, in which Swin Transformer absorbs the local merit of CNN and global advantage to encode hierarchical features, spatial alignment and channel re-calibration module enhances the intra-layer cross-modality features, edge-guided decoder strengths the inter-layer cross-modality fusion. Supervised by edge and saliency map, SwinNet works excellent on public databases. Increasing accuracy also brings about a reduction in speed. In the future, we will discuss the lightweight design.


\end{document}